\documentclass{article} 
\usepackage{iclr2023_conference,times}

\usepackage{amsmath,amsfonts,bm}

\def\eqref#1{equation~\ref{#1}}

\def\1{\bm{1}}

\DeclareMathAlphabet{\mathsfit}{\encodingdefault}{\sfdefault}{m}{sl}
\SetMathAlphabet{\mathsfit}{bold}{\encodingdefault}{\sfdefault}{bx}{n}

\newcommand{\R}{\mathbb{R}}

\usepackage{float}
\usepackage{hyperref}
\usepackage{url}

\usepackage{graphicx}
\usepackage{wrapfig,lipsum,booktabs}
\graphicspath{ {./images/} }
\usepackage{caption}
\usepackage{subcaption}
\usepackage{hhline}
\usepackage{multirow}
\usepackage{algorithm}
\usepackage{algpseudocode}
\usepackage{amsthm}

\usepackage{listings}
\usepackage{xcolor}

\definecolor{codegreen}{rgb}{0,0.6,0}
\definecolor{codegray}{rgb}{0.5,0.5,0.5}
\definecolor{codepurple}{rgb}{0.58,0,0.82}
\definecolor{backcolour}{rgb}{0.95,0.95,0.92}

\lstdefinestyle{mystyle}{
  backgroundcolor=\color{backcolour}, commentstyle=\color{codegreen},
  keywordstyle=\color{magenta},
  numberstyle=\tiny\color{codegray},
  stringstyle=\color{codepurple},
  basicstyle=\ttfamily\footnotesize,
  breakatwhitespace=false,         
  breaklines=true,                 
  captionpos=b,                    
  keepspaces=true,                 
  numbers=left,                    
  numbersep=5pt,                  
  showspaces=false,                
  showstringspaces=false,
  showtabs=false,                  
  tabsize=2
}

\numberwithin{equation}{section}
\newtheorem{theorem}{Theorem}

\title{Random Weight Factorization improves the training of Continuous Neural Representations}

\author{Sifan Wang, \ Hanwen Wang, \ Jacob H. Seidman, \ Paris Perdikaris
\\
University of Pennsylvania, Philadelphia, PA 19104 \\
\texttt{\{sifanw, wangh19, seidj@sas.upenn.edu\}@sas.upenn.edu}, \\ \texttt{pgp@seas.upenn.edu}
}

\iclrfinalcopy 

\begin{document}







%

\maketitle
\begin{abstract}
Continuous neural representations have recently emerged as a powerful and flexible alternative to classical discretized representations of signals. However, training them to capture fine details in multi-scale signals is difficult and computationally expensive. Here we propose {\em random weight factorization} as a simple drop-in replacement for parameterizing and initializing conventional linear layers in coordinate-based multi-layer perceptrons (MLPs) that significantly accelerates and improves their training. We show how this factorization alters the underlying loss landscape and effectively enables each neuron in the network to learn using its own self-adaptive learning rate. This not only helps with mitigating spectral bias, but also allows networks to quickly recover from poor initializations and reach better local minima. We demonstrate how {\em random weight factorization} can be leveraged to improve the training of neural representations on a variety of tasks, including image regression, shape representation, computed tomography, inverse rendering, solving partial differential equations, and learning operators between function spaces.
\end{abstract}

\section{Introduction}


Some of the recent advances in machine learning can be attributed to new developments in the design of  \textit{continuous neural representations}, which employ coordinate-based multi-layer perceptrons (MLPs) to parameterize discrete signals (e.g. images, videos, point clouds) across space and time.  Such parameterizations are appealing because they are differentiable and much more memory efficient than grid-sampled representations,  naturally allowing smooth interpolations to unseen input coordinates. As such, they have achieved widespread success in a variety of computer vision and graphics tasks, including image representation \citep{stanley2007compositional, nguyen2015deep}, shape representation \citep{chen2019learning, park2019deepsdf, genova2019learning, genova2020local}, view synthesis \citep{sitzmann2019scene, saito2019pifu, mildenhall2020nerf, niemeyer2020differentiable}, texture generation \citep{oechsle2019texture, henzler2020learning},  etc. Coordinate-based MLPs have also been applied to scientific computing applications such as physics-informed neural networks (PINNs) for solving forward and inverse partial differential equations (PDEs) \cite{raissi2019physics, raissi2020hidden, karniadakis2021physics}, and Deep Operator networks (DeepONets) for learning operators between infinite-dimensional function spaces \cite{lu2021learning, wang2021learning}.


Despite their flexibility, it has been shown both empirically and theoretically that coordinate-based MLPs suffer from ``spectral bias'' \citep{rahaman2019spectral, cao2019towards, xu2019frequency}. This manifests as a difficulty in learning the high frequency components and fine details of a target function. A popular method to resolve this issue is to embed input coordinates into a higher dimensional space, for example by using Fourier features before the MLP \citep{mildenhall2020nerf, tancik2020fourier}. 
Another widely used approach is the use of SIREN networks \citep{sitzmann2020implicit}, which employs MLPs with periodic activations to represent complex natural signals and their derivatives. One main limitation of these methods is that a number of associated hyper-parameters (e.g. scale factors) need to be carefully tuned in order to avoid  catastrophic generalization/interpolation errors. Unfortunately, the selection of appropriate hyper-parameters typically requires some prior knowledge about the target signals, which may not be available in some applications.


More general approaches to improve the training and performance of MLPs involve different types of normalizations, such as Batch Normalization \citep{ioffe2015batch}, Layer Normalization \citep{ba2016layer} and Weight Normalization \citep{salimans2016weight}. However, despite their remarkable success in deep learning benchmarks, these techniques are not widely used in MLP-based neural representations. Here we draw motivation from the work of \citep{salimans2016weight,wang2021enhancing} and investigate a simple yet remarkably effective re-parameterization of  weight vectors in MLP networks, coined as \textit{random weight factorization}, which provides a generalization of Weight Normalization and demonstrates significant performance gains. Our main contributions are summarized as
\begin{itemize}
    \item We show that \textit{random weight factorization} alters the loss landscape of a neural representation in a way that can drastically reduce the distance between different parameter configurations, and effectively assigns a self-adaptive learning rate to each neuron in the network. 
    
    \item We empirically illustrate that \textit{random weight factorization} can effectively mitigate spectral bias, as well as enable coordinate-based MLP networks to escape from poor intializations and find better local minima.
    
    \item We demonstrate that \textit{random weight factorization} can be used as a simple drop-in enhancement to conventional linear layers, and yield consistent and robust improvements across a wide range of tasks in computer vision, graphics and scientific computing.
\end{itemize}



\section{Weight Factorization}
Let $\bm{x} \in \R^d $ be the input, $\bm{g}^{(0)}(\bm{x}) = \bm{x}$ and $d_0 = d$. We consider a standard multi-layer perceptron (MLP)
$f_{\bm{\theta}}(\bm{x})$ recursively defined by
\begin{align}
    \label{eq: mlp_1}
    \bm{f}_\theta^{(l)}(\bm{x}) = \bm{W}^{(l)} \cdot \bm{g}^{(l-1)}(\bm{x}) + \bm{b}^{(l)}, \quad \bm{g}^{(l)}(\bm{x}) = \sigma(\bm{f}_\theta^{(l)}(\bm{x})), \quad l = 1,2, \dots, L,
\end{align}
with a final layer 
\begin{align}
    \label{eq: mlp_2}
    f_{\bm{\theta}}(\bm{x}) &= \bm{W}^{(L+1)} \cdot \bm{g}^{(L)}(\bm{x}) + \bm{b}^{(L+1)},
\end{align}
where $\bm{W}^{(l)} \in \R^{d_l \times d_{l-1}}$ is the weight matrix in $l$-th layer and $\sigma$ is an element-wise activation function. Here, $\bm{\theta}=\left(\bm{W}^{(1)}, \bm{b}^{(1)}, \ldots, \bm{W}^{(L+1)},  \bm{b}^{(L+1)}\right)$ 
represents all trainable parameters in the network. 

MLPs are commonly trained by minimizing an appropriate  loss function $\mathcal{L}(\bm{\theta})$ via gradient descent. To improve convergence, we propose to factorize the weight parameters associated with each neuron in the network as follows
\begin{align}
    \label{eq: neuron_fact}
    \bm{w}^{(k, l)} =  s^{(k, l)}  \cdot \bm{v}^{(k, l)}, \quad k=1, 2, \dots, d_l, \quad l = 1,2, \dots, L+1,
\end{align}
where $\bm{w}^{(k,l)} \in \R^{d_{l-1}}$ is a weight vector representing the $k$-th row of the weight matrix $\bm{W}^{(l)}$, $s^{(k,l)} \in \R$ is a trainable scale factor assigned to each individual neuron, and $\bm{v}^{(k,l)} \in \R^{d_{l-1}}$. Consequently, the proposed weight factorization can be written by
\begin{align}
    \label{eq: weight_fact}
    \bm{W}^{(l)} = \mathrm{diag}(\bm{s}^{(l)}) \cdot  \bm{V}^{(l)}, \quad l = 1,2, \dots, L+1.
\end{align}
with $\bm{s} \in \R^{d_l}$.

\subsection{A Geometric Perspective}
\label{sec: geo}

In this section, we provide a geometric motivation for the proposed weight factorization. To this end, we consider the simplest setting of a one-parameter loss function $\ell(w)$. For this case, the weight factorization is reduced to $w = s \cdot v$ with two scalars $s, v$. Note that for a given $w \neq 0$ there are infinitely many pairs $(s,v)$ such that $w=s \cdot v$. The set of such pairs forms a family of hyperbolas in the $sv$-plane (one for each choice of signs for both $s$ and $v$). As such, the loss function in the $sv$-plane is constant along these hyperbolas.

\begin{wrapfigure}[13]{r}{0.7\textwidth}
\vspace{-2mm}
  \begin{center}
    \includegraphics[width=0.7\textwidth]{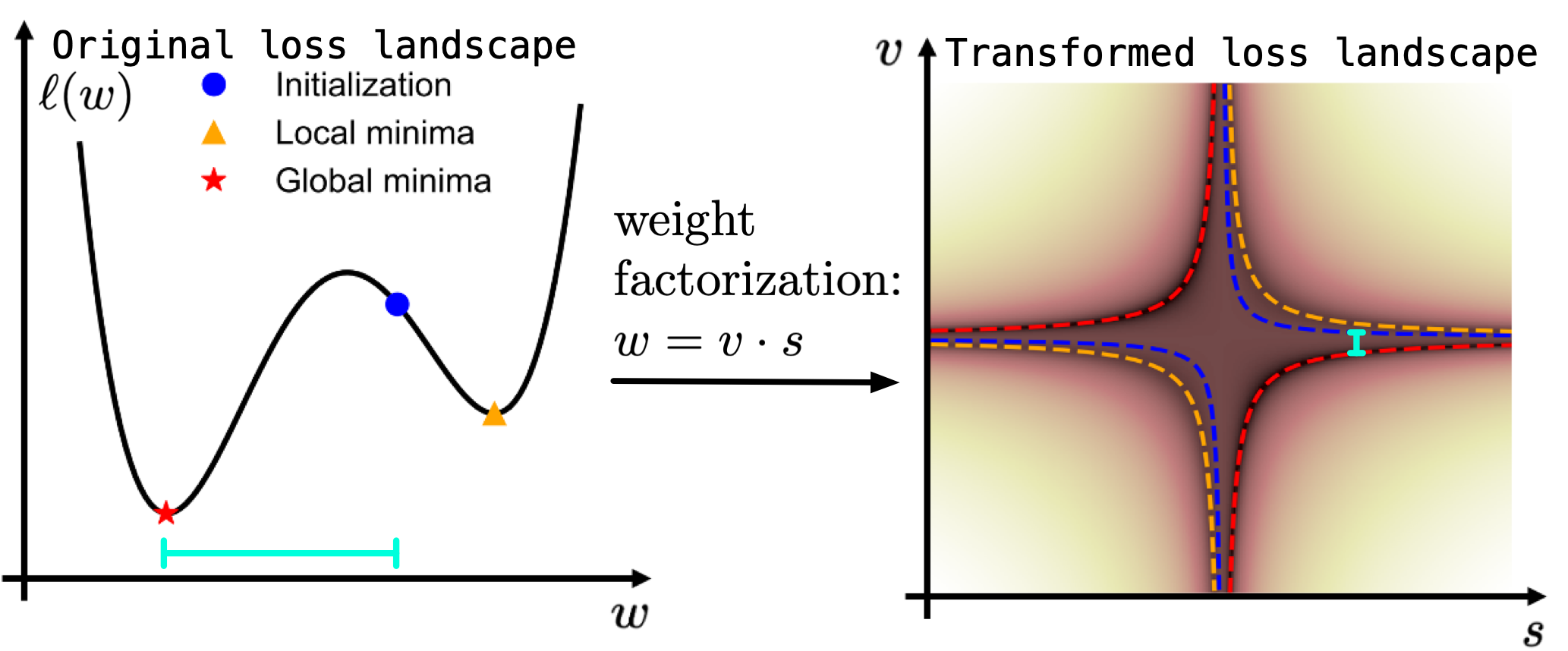}
  \end{center}
  \vspace{-5mm}
  \caption{{ Weight factorization transforms loss landscapes and shortens the distance to minima.} }
  \label{fig: loss_landscape}
\end{wrapfigure}

Figure \ref{fig: loss_landscape} gives a visual illustration of the difference between the original loss landscape as a function of $w$ versus the loss landscape in the factorized $sv$-plane.  In the left panel, we plot the original loss function as well as an initial parameter point, the local minimum, and the global minimum. The right panel shows how in the factorized parameter space, each of these three points corresponds to two hyperbolas in the $sv$-plane. Note how the distance between the initialization and the global minima is reduced from the top to the bottom panel upon an appropriate choice of factorization. The key observation is that the distance between factorizations representing the initial parameter and the global minimum become arbitrarily small in the $sv$-plane for larger values of $s$. Indeed, we can prove that this holds for any general loss function in arbitrary parameter dimensions (the proof is provided in Appendix \ref{appendix: thm1}).

\begin{theorem}
Suppose that $\mathcal{L}(\bm{\theta})$ is the associated loss function of a neural network defined in \eqref{eq: mlp_1} and \eqref{eq: mlp_2}. For a given $\bm{\theta}$, we define $U_{\bm{\theta}}$ as the set containing all possible weight factorizations 
\begin{align}
    U_{\bm{\theta}} = \left\{ (\bm{s}^{(l)}, \bm{V}^{(l)})_{l=1}^{L+1}  : \mathrm{diag}(\bm{s}^{(l)}) \cdot  \bm{V}^{(l)}   = \bm{W}^{(l)}, \quad l=1, \dots, L+1  \right\}.
\end{align}
Then for any $\bm{\theta}, \bm{\theta}'$, we have
\begin{align}
    \text{dist}(U_{\bm{\theta}}, U_{\bm{\theta}'}) = 0.
\end{align}
\end{theorem}

\subsection{Self-adaptive learning rate for each neuron}

A different way to examine the effect of the proposed weight factorization is by studying its associated gradient updates. Recall that a standard gradient descent update with a learning rate $\eta$ takes the form 
\begin{align}  
    \label{eq: gradient_update}
    \bm{w}^{(k,l)}_{n+1} &= \bm{w}^{(k,l)}_{n} - \eta \frac{\partial \mathcal{L}}{\partial \bm{w}^{(k,l)}_{n}}.
\end{align}
The following theorem derives the corresponding gradient descent update expressed in the original parameter space for models using the proposed weight factorization.
\begin{theorem} Under the weight factorization of \eqref{eq: neuron_fact}, the gradient descent update is given by
\begin{align}
    \label{eq: fact_gradient_update}
    \bm{w}^{(k, l)}_{n+1} = \bm{w}^{(k, l)}_{n} - \eta \left(\| [s^{(k,l)}_n]^2 + \bm{v}^{(k,l)}_n\|^2_2 \right)  \frac{\partial \mathcal{L}}{\partial \bm{w}^{(k,l)}_{n}}  + \mathcal{O}(\eta^2),
\end{align}
for $l=1, 2,\dots, L+1$ and $k=1, 2, \dots, d_l$.
\end{theorem}

The proof is  provided in Appendix \ref{appendix: thm2}.  By comparing \eqref{eq: gradient_update} and  \eqref{eq: fact_gradient_update}, we observe that the weight factorization $\bm{w} = s \cdot \bm{v}  $ re-scales the learning rate of $\bm{w}$ by a factor of $(s^2 + \|\bm{v}\|_2^2)$. Since $\bm{s}, \bm{v}$ are trainable parameters, this analysis suggests that this weight factorization effectively assigns a self-adaptive learning rate to each neuron in the network. In the following sections, we will demonstrate that the proposed weight factorization (under an appropriate initialization of the scale factors), not only helps with mitigating spectral bias \citep{rahaman2019spectral,bietti2019inductive,tancik2020fourier,wang2021eigenvector}, but also allows networks to quickly move away from a poor initialization and reach better local minima faster.

\subsection{Relation to existing works} 
\label{sec: A.A&W.N}

The proposed weight factorization is largely motivated by \textit{weight normalization} \citep{salimans2016weight}, which decouples the norm and the directions of the weights associated with each neuron as
\begin{align}
   \label{eq: weight_norm}
    \bm{w} = g \frac{\bm{v}}{\|\bm{v}\|},
\end{align}
where $g = \|\bm{w}\|$, and gradient descent updates are applied directly to the new parameters $\bm{v}, g$. Indeed, this can be viewed as a special case of the proposed weight factorization by setting $\bm{s} = \|\bm{w}\|$ in \eqref{eq: neuron_fact}. In contrast to weight normalization, our weight factorization scheme allows for more flexibility in the choice of the scale factors $\bm{s}$. 

We note that SIREN networks \citep{sitzmann2020implicit} also employ a special weight factorization for each   hidden layer weight matrix, 
\begin{align}
    \bm{W}=\omega_0 * \hat{\bm{W}},
\end{align}
where the scale factor $\omega_0 \in \R$ is a user-defined hyper-parameter.  Although the authors attribute the success of SIREN to the periodic activation functions in conjunction with a tailored initialization scheme, here we will demonstrate that  the specific choice of $\omega_0$ is the most crucial element in SIREN's performance, see Appendix \ref{appendix: 2d_image}, \ref{appendix: CT} for more details.

It is worth pointing out that the proposed weight factorization also bears some resemblance to the adaptive activation functions introduced in  \citep{jagtap2020adaptive}, which modifies the activation of each neuron by introducing  an additional trainable parameter $\bm{a}$ as
\begin{align}
    \bm{g}^{(l)}(\bm{x}) = \sigma (\bm{a} \bm{f}^{(l)}(\bm{x})).
\end{align} 
These adaptive activations aim to help networks learn sharp gradients and transitions of the target functions.  In practice, the scale factor is generally initialized as $\bm{a}=\bm{1}$, yielding a trivial weight factorization. As illustrated in the next section, this is fundamentally different from our approach as we initialize the scale factors $\bm{s}$ by a random distribution and re-parameterize the weight matrix accordingly. In Section \ref{sec: results} we demonstrate that, by initializing $\bm{s}$ using an appropriate distribution, we can consistently outperform both weight normalization, SIREN, and adaptive activations across a broad range of supervised and self-supervised learning tasks.

\section{Random weight Factorization in practice}
\label{sec: rwf_practice}


Here we illustrate the use of weight factorization through the lens of a simple regression task. Specifically, we consider a smooth scalar-valued function $f$ sampled from a Gaussian random field using a square exponential kernel with a length scale of $l=0.02$. This generates a data-set of $N=256$ observation pairs $\{x_i,f(x)_i\}_{i=1}^N$, where $\{x_i\}_{i=1}^N$ lie on a uniform grid in $[0, 1]$.  The goal is to train a network $f_{\bm{\theta}}$ to learn $f$ by minimizing the mean square error loss $\mathcal{L}(\bm{\theta}) = 1/N \sum_{i=1}^N |f_{\bm{\theta}}(x_i) - f(x_i)|^2$. 

The proposed random weight factorization is applied as follows. We first initialize the parameters of an MLP network via the Glorot scheme \citep{glorot2010understanding}. Then, for every weight matrix $\bm{W}$, we proceed by initializing a scale vector $\exp(\bm{s})$ where $\bm{s}$ is sampled from a multivariate normal distribution $\mathcal{N}(\bm{\mu}, \sigma \mathrm{I})$. Finally, every weight matrix is factorized by the associated scale factor as  $ \bm{W} = \mathrm{diag}(\exp(\bm{s})) \cdot \bm{V}$ at initialization.  We train this network by gradient descent on the new parameters $\bm{s}, \bm{V}$ directly. This procedure is summarized in Appendix  \ref{appendix: alg}, along with a simple JAX Flax implementation  \citep{flax2020github} in Appendix \ref{appendix: flax}.

In Figure \ref{fig: 1d_regression}, we train networks (3 layers, 128 neurons per layer, ReLU activations)  to learn the target function using: (a) a conventional MLP, (b) an MLP with adaptive activations (AA) \citep{jagtap2020adaptive}, (c) an MLP with weight normalization (WN) \citep{salimans2016weight}, and (d) an MLP with the proposed random weight factorization scheme (RWF). Evidently, RWF yields the best predictive accuracy and loss convergence. Moreover, we plot the relative change of the weights in the original (unfactorized) parameter space during training in the bottom middle panel. We observe that RWF leads to the largest weight change during training, thereby enabling the network to find better local minima further away from its initialization. To further emphasize the benefit of weight factorization, we compute the eigenvalues of the resulting empirical Neural Tangent Kernel (NTK) \citep{jacot2018neural}
\begin{align}
    \bm{K}_{\bm{\theta}} = \left<\frac{\partial f_{\bm{\theta}}}{\partial \bm{\theta}}(x_i),  \frac{\partial f_{\bm{\theta}}}{\partial \bm{\theta}}(x_j) \right>_{ij},
\end{align}
at the last step of training and visualize them in the bottom right panel.  Notice how RWF exhibits a flatter NTK spectrum and slower eigenvalue decay than the other methods, indicating better-conditioned training dynamics and less severe spectral bias, see  \citep{rahaman2019spectral,bietti2019inductive,tancik2020fourier,wang2021eigenvector} for more details. To explore the robustness of the proposed RWF, we conduct a systematic study on the effect of $\mu$ and $\sigma$ in the initialization of the scale factor $\bm{s}$. The results suggest that the choice of $\mu, \sigma$ plays an important role.  Specifically,  too small $\mu, \sigma$ values may lead to performance that is similar to a conventional MLP, while setting $\mu, \sigma$ too large can result in an unstable training process.  We empirically find that $\mu=1, \sigma=0.1$  consistently improves the loss convergence and model accuracy  for  the vast majority of tasks considered in this work. Additional details are presented in Appendix \ref{appendix: 1d_regression}.

\begin{figure}
     \centering
     \begin{subfigure}[b]{0.9\textwidth}
         \centering
         \includegraphics[width=\textwidth]{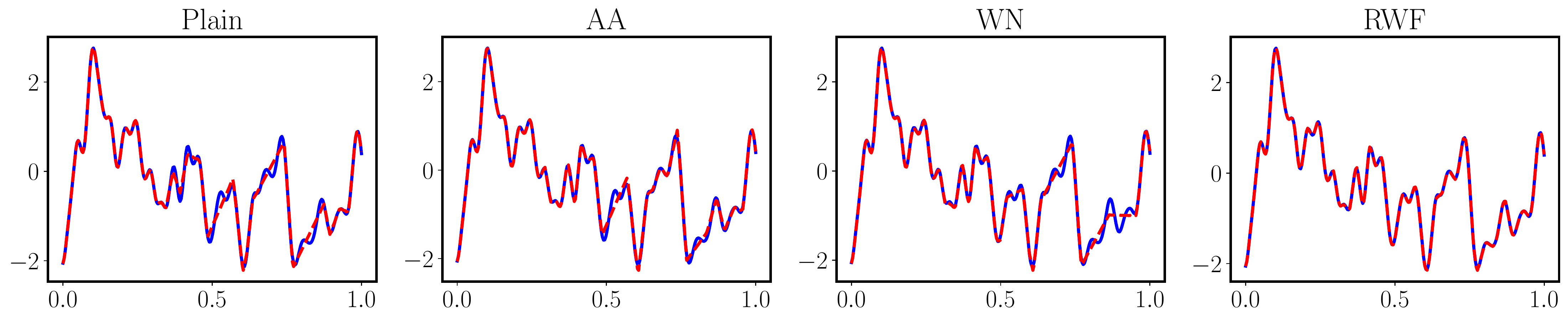}
         \label{fig: 1d_regression_preds}
     \end{subfigure}
     \hfill
     \begin{subfigure}[b]{0.9\textwidth}
         \centering
         \includegraphics[width=\textwidth]{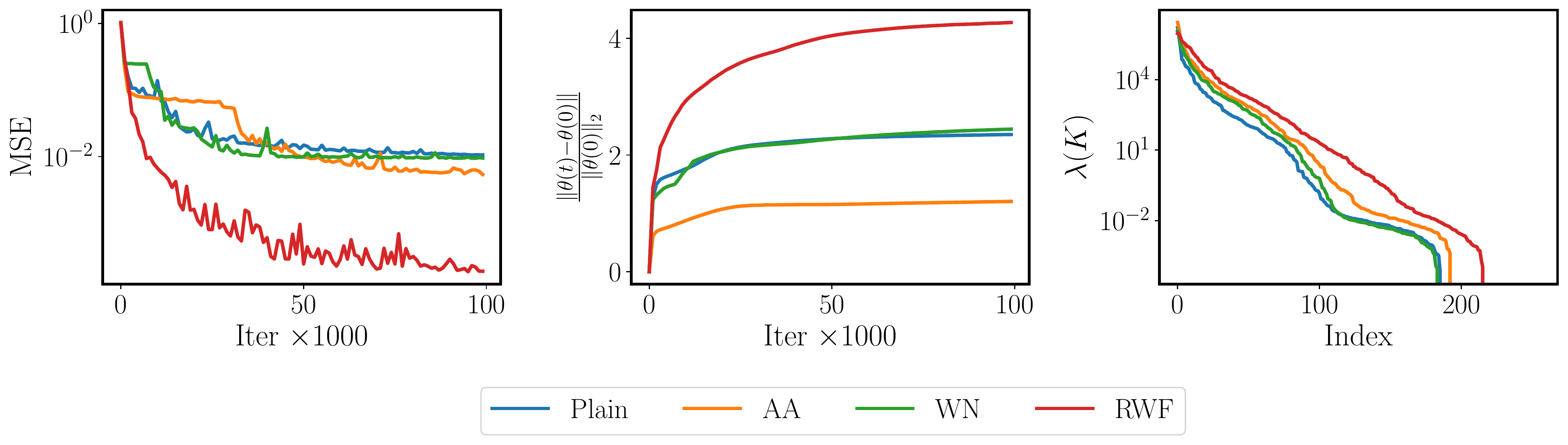}
         \label{fig: 1d_regression_stats}
     \end{subfigure}
     \vspace{-5mm}
        \caption{{\em 1D regression:} {\em Top:} Model predictions using different parameterizations.  Plain: Standard MLP; AA: adaptive activation; WN: weight normalization; RWF: random weight factorization. {\em Bottom left:} Mean square error (MSE)  during training. {\em Bottom Middle:} Relative change of weights during training. The comparison is performed in the original parameter space. {\em Bottom right:} Eigenvalues (descending order) of the empirical NTK at the end of training.  }
        \label{fig: 1d_regression}
\end{figure}

\section{Experiments}
\label{sec: results}
In this section, we demonstrate the effectiveness and robustness of {\em random weight factorization} for training continuous neural representations across a range of tasks in computer vision, graphics, and  scientific computing. More precisely, we compare the performance of plain MLPs, MLPs with adaptive activations (AA) \citep{jagtap2020adaptive}, weight normalization (WN) \citep{salimans2016weight}, and the proposed random weight factorization (RWF). The comparison is performed over a collection of MLP architectures, including conventional MLPs, SIREN \citep{sitzmann2020implicit}, modified MLPs \citep{wang2021understanding}, as well as
MLPs with positional encodings \citep{mildenhall2020nerf} and Gaussian Fourier features \citep{tancik2020fourier}. The hyper-parameters of our experiments along with the computational cost associated with each experiment are presented in Appendix \ref{appendix: hyper_parameters} and Appendix \ref{appendix: computational_cost}, respectively. Notice that the computational overhead of our method is marginal, and RWF can be therefore considered as a drop-in enhancement to any architecture that uses linear layers. Table \ref{tab: results} summarizes the results obtained for each benchmark, corresponding to the optimal input mapping and network architecture.  Overall, RWF consistently achieves the best performance across tasks and architectures. All code and data will be made publicly available. A summary of each benchmark study is presented below, with more details provided in Appendix.

\begin{table}
    \centering
      \renewcommand{\arraystretch}{1.2}
      \resizebox{\textwidth}{!}
     {
    \begin{tabular}{c|c|c|c|c|c|c}
     Task & Metric & Case & Plain  & AA & WN & RWF (ours) \\
     \hline
     \hline
     \multicolumn{1}{c|}{\multirow{2}{*}{Image Regression}} &
     \multicolumn{1}{c|}{\multirow{2}{*}{PSNR ($\uparrow$)}} &  Natural &  27.35 &   27.37& 27.36 & \textbf{28.08}  \\
                                     &   & Text  &  32.09 &   32.29& 32.25 & \textbf{33.13} \\
    \hline
     \multicolumn{1}{c|}{\multirow{3}{*}{Shape Representation}} &
     \multicolumn{1}{c|}{\multirow{2}{*}{IoU ($\uparrow$)}} &  Dragon & 0.980  & 0.981 & 0.981 & \textbf{0.984}\\
                            &  & Armadillo & 0.978  & 0.976 & 0.974 & \textbf{0.982}\\
    \hline
     \multicolumn{1}{c|}{\multirow{2}{*}{Computed Tomography}} &
     \multicolumn{1}{c|}{\multirow{2}{*}{PSNR ($\uparrow$)}} &  Shepp & 30.09 & 30.37 & 30.59  & \textbf{33.73}  \\
                         &   & ATLAS  & 21.85 & 21.93    & 22.16 & \textbf{23.71} \\
    \hline
    Inverse Rendering  & PSNR ($\uparrow$) &Lego  & 25.99 & 25.93  & 25.93 & \textbf{26.13}\\
    \hline
    \multicolumn{1}{c|}{\multirow{2}{*}{Solving PDEs}} &
     \multicolumn{1}{c|}{\multirow{2}{*}{Rel. $L^2$ ($\downarrow$)}} &  Advection &   28.82\% & 38.63\% & 46.34\% &  \textbf{4.14\%} \\
                          &   & Navier-Stokes & 39.25\%  &  34.09\% & 30.98\% & \textbf{6.67\%} \\
    \hline
     \multicolumn{1}{c|}{\multirow{3}{*}{Learning Operators}} &
     \multicolumn{1}{c|}{\multirow{3}{*}{Rel. $L^2$ ($\downarrow$)}} &  DR & 1.09\% &  0.95\% & 0.97\% & \textbf{0.50\%}  \\
                                                                     &   & Darcy & 2.03\% &  2.06\% & 2.00\% & \textbf{1.67\%}  \\
      &   & Burgers &  5.11\%  &  4.71\%  &   4.37\% &  \textbf{2.46\%}  \\
    \hline
    \end{tabular}
    }
    \caption{We compare four different parameterizations over various benchmarks, and demonstrate that \textit{Random weight factorization} consistently outperforms other parameterizations  across all tasks. All comparisons are conducted under exactly the same hyper-parameter settings. $(\uparrow)$/$(\downarrow)$ indicates that higher/lower values are better, respectively. (Plain: conventional MLP; AA: adatpive activation; WN: weight normalization; RWF: random weight factorization). }
    \label{tab: results}
\end{table}

\subsection{2D image regression} 
\label{sec: 2d_img}

We train coordinate-based MLPs to learn a map from 2D input pixel coordinates to the corresponding RGB values of an image, using the benchmarks put forth in  \citep{tancik2020fourier}. We conduct experiments using two data-sets: \textit{Natural} and \textit{Text}, each containing 16 images. The \textit{Natural} data-set is constructed by taking center crops of randomly sampled images from the Div2K data-set \citep{agustsson2017ntire}. The \textit{Text} data-set is constructed by placing random text with random font sizes and colors on a white background. The training data is obtained by downsampling each test image by a factor of 2.  We compare the resulting peak signal-to-noise ratio (PSNR) in the full resolution test images, obtained with different MLP architectures using different input mappings and weight parametrizations.

\subsection{3D shape representation} 
\label{sec: 3d_shape}

This task follows the original problem setup in \citep{tancik2020fourier}. The goal is to learn an implicit representation of a 3D shape using Occupancy networks \citep{mescheder2019occupancy}, which take spatial coordinates as inputs and predict $0$ for points outside a given shape and $1$ for points inside the shape. We use two complex triangle meshes commonly used in computer graphics: \textit{Dragon} and  \textit{Armadillo}. The training data is generated by randomly sampling points inside a bounding box and calculating their labels using the ground truth mesh. We evaluate the trained model performance using the Intersection over Union (IoU) metric on a set of points randomly sampled near the mesh surface to better highlight the different mappings’ abilities to resolve fine details.

\subsection{2D computed tomography (CT)} 
\label{sec: 2d_ct}

This task follows the original problem setup in \citep{tancik2020fourier}. We train an MLP to learn a map from 2D pixel coordinates to a corresponding volume density at those locations. Two data-sets are considered: procedurally-generated 
Shepp-Logan phantoms \citep{shepp1974fourier} and 2D brain images from the ATLAS data-set  \citep{liew2018large}. Different from the previous tasks, we observe integral projections of a density field instead of direct measurements. The network is trained in an indirect supervised fashion by minimizing a loss between a sparse set of ground-truth integral projections and integral projections computed from the network’s output. We use PSNR to quantify the performance of the trained MLPs with different input mappings and different weight parametrizations.

\subsection{3D inverse rendering for view synthesis} 
\label{sec: nerf}
This task follows the original problem setup in \citep{tancik2020fourier}. We aim to learn an implicit representation of a 3D scene from 2D photographs using Neural Radiance Field (NeRF) \citep{mildenhall2020nerf}, which is a coordinate-based MLP that takes a 3D location as input and outputs a color and volume density. The network is trained by minimizing a rendering loss between the set of 2D image observations and the same rendered views from the predicted scene representation. In our experiments, we consider a down-sampled NeRF \textit{Lego} data-set and use a simplified version
of the method described in \citep{mildenhall2020nerf}, where we remove hierarchical sampling and view dependence. We compare the PSNR of the trained MLPs with different weight parametrizations.

\subsection{Solving partial differential equations (PDEs)}
\label{sec: pinns}

Our goal is to solve partial differential equations (PDEs) using physics-informed neural network (PINNs) \citep{raissi2019physics}, which take the coordinates of a spatio-temporal domain as inputs and predict the corresponding target solution function.  PINNs are trained in a self-supervised fashion by minimizing a composite loss function for fitting given initial and boundary conditions, as well as satisfying the underlying PDE constraints. We consider two benchmarks, an advection equation modeling the transport of a scalar field, and the Navier-Stokes equation modeling the motion of an incompressible fluid in a square cavity. Detailed descriptions and implementations of each problem are provided below and in the Appendix \ref{appendix: pinns}. 

\begin{figure}
    \centering
    \includegraphics[width=0.9\textwidth]{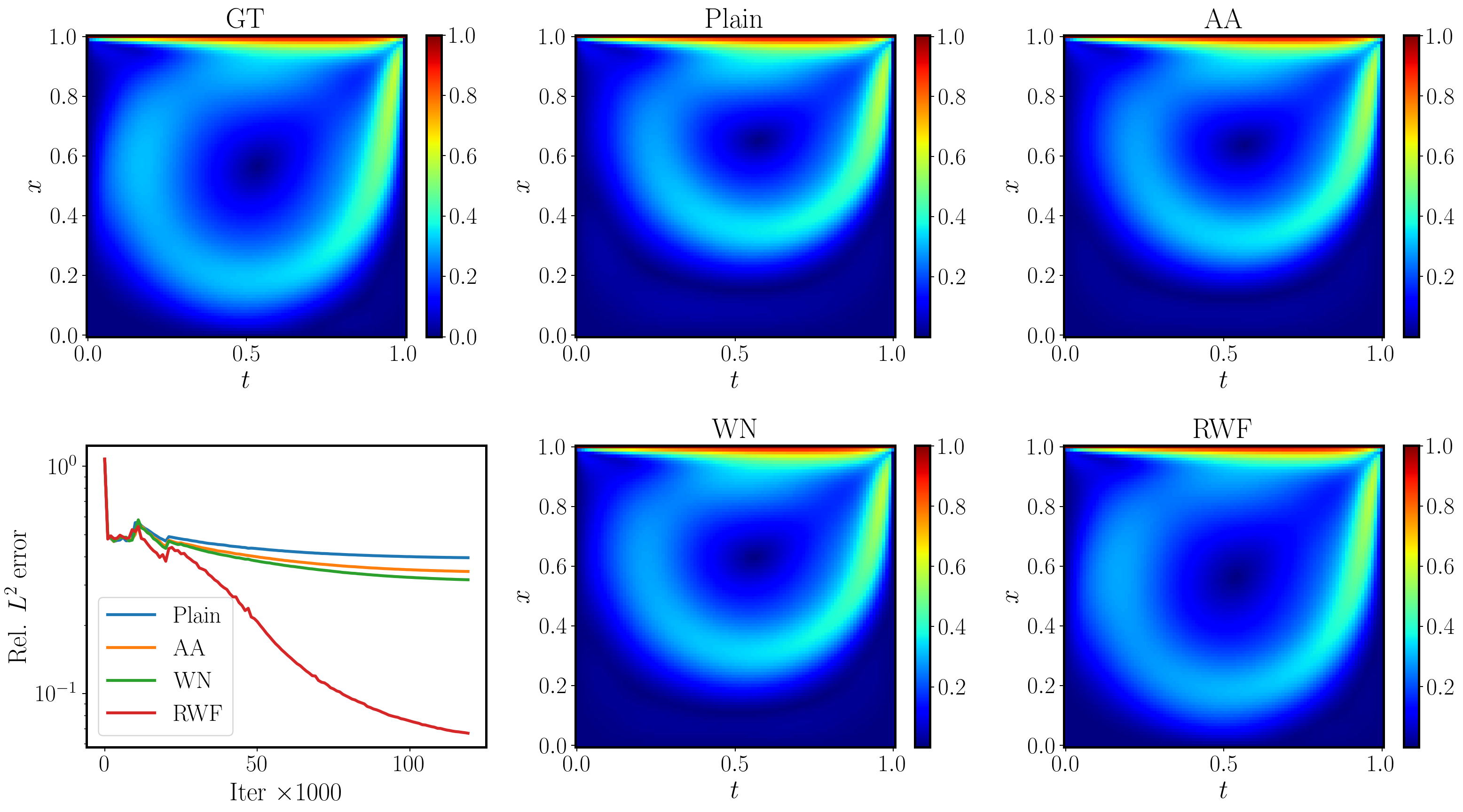}
    \caption{{\em Navier-Stokes:} Predicted solutions of trained MLPs with different weight parameterizations, along with the evolution of the associated relative $L^2$ errors during training.}
    \label{fig: pinn_ns}
\end{figure}

Figure \ref{fig: pinn_adv} and Figure \ref{fig: pinn_ns} present the ground truth against the predicted solutions obtained by training PINNs with different weight parameterizations.  It can be observed that the predictions obtained by RWF are in excellent agreement with the ground truth, while the other three parameterizations result in poor or even non-physical approximations. The rapid decrease in the test error further validates the benefit of our method. We attribute these significant performance improvements to the fact that PINN models, due to their self-supervised nature, often suffer from poor initializations. Evidently, RWF can precisely mitigate this by being able to reach better local minima that are located further away from the model initialization neighborhood.


\paragraph{Advection equation:} The first example is 1D advection equation,  a linear hyperbolic equation commonly used to model transport phenomena
\begin{align}
    \frac{\partial u}{\partial t}+ c \frac{\partial u}{\partial x} &=0, \quad x \in (0, 2 \pi), t \in[0, 1], \\
    u(x, 0) &=g(x), \quad x \in (0, 2 \pi),
\end{align}
with periodic boundary conditions. This example has been studied in \citep{krishnapriyan2021characterizing, daw2022rethinking}, exposing some of the limitations that PINNs suffer from as the transport velocity $c$ is increased. In our experiments, we consider $c=50$ and an initial condition $g(x) = \sin(x)$.

\begin{figure}
    \centering
    \includegraphics[width=0.9\textwidth]{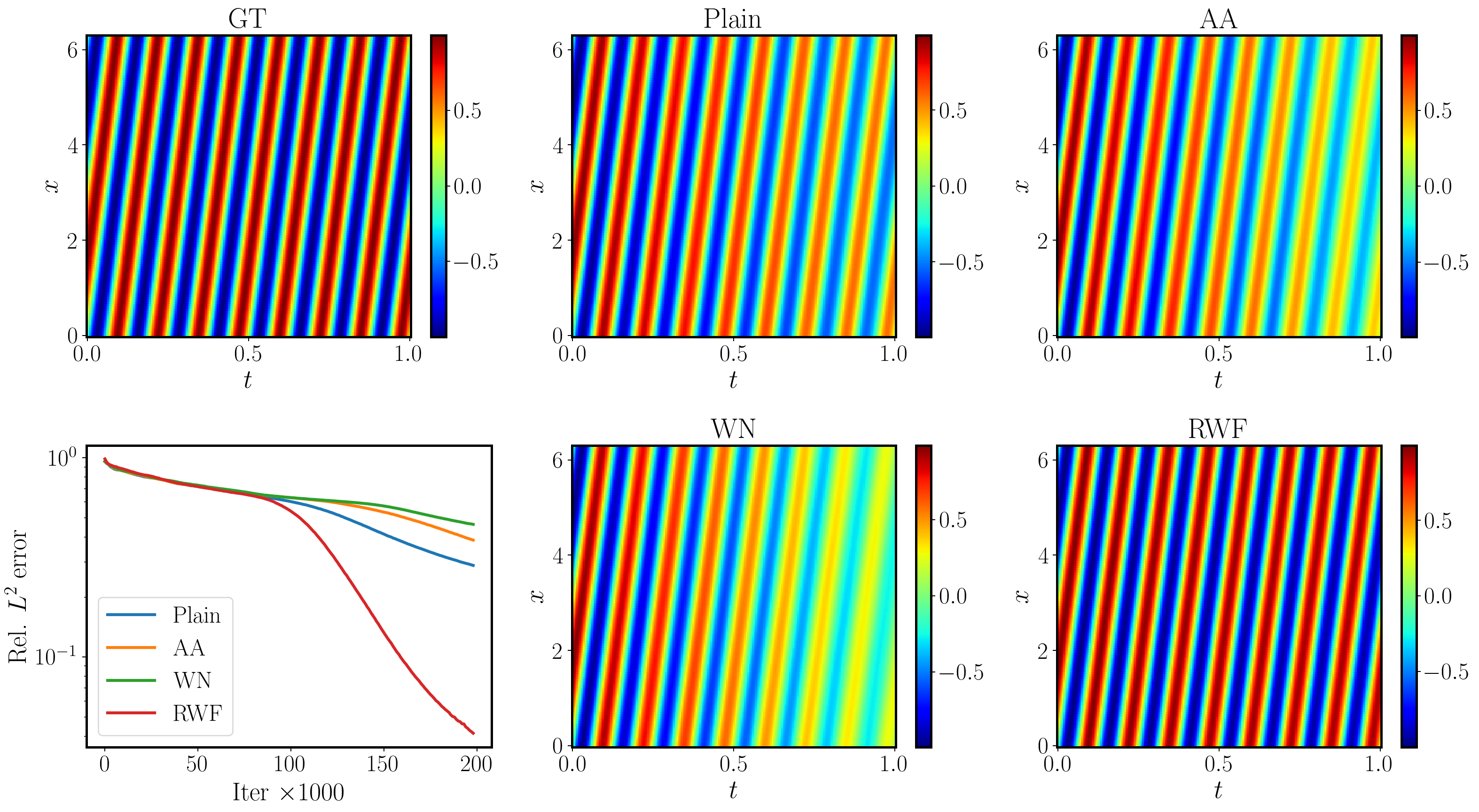}
    \caption{{\em Adection:} Predicted solutions of trained MLPs with different weight parameterizations, along with the evolution of the associated relative $L^2$ prediction errors during training.}
    \label{fig: pinn_adv}
\end{figure}


\paragraph{Navier-Stokes equation:} The second example is a classical benchmark problem in computational fluid dynamics, describing the motion of an incompressible fluid in a two-dimensional lid-driven cavity. The system is governed by the Navier–Stokes equations written in a non-dimensional form
\begin{align}
    \bm{u} \cdot \nabla \bm{u}+\nabla p-\frac{1}{R e} \Delta \bm{u}&=0, \quad  (x,y) \in (0,1)^2, \\
    \nabla \cdot \bm{u}&=0, \quad  (x,y) \in (0,1)^2,
\end{align}
where $\bm{u} = (u, v)$ denotes the velocity in $x$ and $y$ directions, respectively, and $p$ is the scalar pressure field. We assume $\bm{u}=(1, 0)$ on the top lid of the cavity, and a non-slip boundary condition on the other three walls. All experiments are performed with a Reynolds  number of $Re =1,000$.

\subsection{Learning Operators}
\label{sec: deeponets}

In this task, we focus on learning the solution operators of parametric PDEs. To describe the problem setup in general, consider a parametric PDE of the following  form
\begin{align}
    \label{eq: parametric_PDE}
    \mathcal{N}(\bm{a}, \bm{u}) = 0,
\end{align}
where  $\mathcal{N}: \mathcal{A} \times \mathcal{U} \rightarrow \mathcal{V}$ is a linear or nonlinear differential operator between infinite-dimensional function spaces. Moreover, $\bm{a} \in \mathcal{A}$ denotes the PDE parameters, and $\bm{u} \in \mathcal{U}$ is the corresponding unknown solutions of the PDE system. The solution operator $G: \mathcal{A} \rightarrow \mathcal{U}$ is given by
\begin{align}
    G(\bm{a}) = \bm{u}(\bm{a}).
\end{align}
In our experiments, we consider three benchmarks: Diffusion-reaction, Darcy flow and the Burgers' equation. Detailed descriptions of each problem setup are shown below.  As shown in  Figure \ref{fig: deeponet_losses}, we plot the training losses of each model with different weight parameterizations. One can see that \textit{random weight factorization} yields the best loss convergence for every example, indicating the capability of the proposed method to accelerate the convergence of stochastic gradient descent and achieve better local minima.

\paragraph{Diffusion-reaction:} Our first example involves  a nonlinear diffusion-reaction PDE with a source term $a: (0, 1) \rightarrow \R$,
\begin{align}
    \frac{\partial u}{\partial t}= D \frac{\partial^{2} u}{\partial x^{2}}+k u^{2}+a(x), \quad (x,t) \in (0,1) \times (0,1],
\end{align}
with the zero initial and boundary conditions, where $D=0.01$ is the diffusion coefficient and $k=0.01$ is the reaction rate. We train a Deep Operator Network (DeepONet)  \citet{lu2021learning} to learn the solution operator for mapping source terms $a(x)$ to the corresponding PDE solutions $u(x)$. This network takes a PDE parameter and a spatial-temporal coordinate as inputs, and predicts the associated PDE solution evaluated at that location. The model is trained in a supervised manner by minimizing a loss between the predicted PDE solutions and the available solution measurements.

\paragraph{Darcy flow:} The Darcy equation describes steady-state flow through a porous medium, taking the following form in two spatial dimensions
\begin{align}
-\nabla \cdot(a \cdot \nabla u) &=f, \quad (x,y) \in(0,1)^{2}, \\
u &=0, \quad    (x,y) \in \partial(0,1)^{2},
\end{align}
where $a: (0, 1) \rightarrow \R^{+}$ is  the diffusion coefficient and $f: (0, 1) \rightarrow \R$ is a forcing term. This is a linear second-order elliptic PDE with numerous applications in modeling subsurface flow, porous media, elastic materials, etc. We fix $f(x,y) = 1$ and aim to learn a continuous representation of the solution operator  $G: a(x,y) \rightarrow u(x,y)$ with a DeepONet. 

\paragraph{Burgers' equation:} As the last example,  we consider a fundamental nonlinear PDE,  the one-dimensional viscous Burgers’ equation. This equation arises in various areas of applied mathematics, such as gas dynamics, nonlinear acoustics, and fluid mechanics. It takes the form
\begin{align}
    \frac{\partial u}{\partial t} + u \frac{\partial u}{\partial x} &= \nu \frac{\partial^2 u}{\partial x^2}, \quad \quad (x,t) \in (0,1) \times (0,1], \\
    u(x, 0)&=0, \quad  x \in (0,1), \\
\end{align}
with periodic boundary conditions and $\nu = 0.001$. Our goal is to learn the solution operator from the initial condition to the associated PDE solution with a physics-informed DeepONet \citep{wang2021learning}. Different from the first two examples,  the physics-informed DeepONet is trained in a self-supervised manner, i.e. without any paired input-output observations, except for a set of given initial or boundary conditions (see \cite{wang2021learning} for more details).

\begin{figure}
    \centering
    \includegraphics[width=0.9\textwidth]{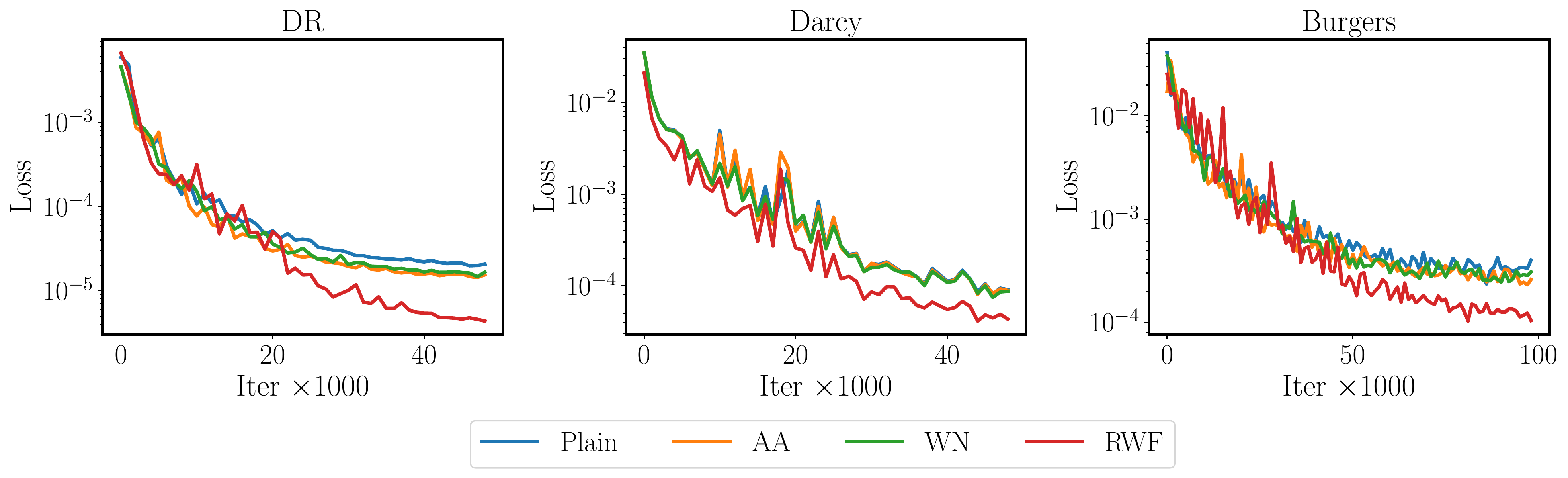}
    \caption{{\em Learning operators:} Loss convergence of training DeepONets with different weight parameterizations for diffusion-reaction equation, Darcy flow and the Burgers' equation.}
    \label{fig: deeponet_losses}
\end{figure}

\section{Conclusions}
In this work, we proposed \textit{random weight factorization}, a simple and remarkably effective  re-parameterization of the weight matrices in neural networks. Theoretically, we show how this factorization alters the geometry of a loss landscape by assigning a self-adaptive learning rate to each neuron. Empirically, we show that our method can mitigate spectral bias in MLPs and enable networks to search for good local optima  further away from their initialization.  We validate \textit{random weight factorization} using six different benchmarks ranging from image regression to learning operators, showcasing a consistent and robust improvements across various tasks in computer vision, graphics, and scientific computing. These findings provide new insights into the training of continuous neural representations and open several exciting avenues for future work, including the application of our method to deep learning models beyond coordinate-based MLPs, such as convolutional networks \citep{lecun1998gradient}, graph networks  \citep{scarselli2008graph}, and Transformers \citep{vaswani2017attention}.

\clearpage

\section*{Author Contributions}
SW and PP conceptualized the research and designed the numerical studies. 
SW, JS and PP provided the theoretical analysis. SW and HW implemented the methods and conducted the numerical experiments. 
PP provided funding and supervised all aspects of this work. All authors discussed the results and contributed to the final manuscript.

\section*{Acknowledgments}
We would like to acknowledge support from the US Department of Energy under the Advanced Scientific Computing Research program (grant DE-SC0019116), the US Air Force (grant AFOSR FA9550-20-1-0060), and US Department of Energy/Advanced Research Projects Agency (grant DE-AR0001201). We also thank the developers of the software that enabled our research, including JAX \citep{jax2018github}, JAX-Flax\citep{flax2020github}, Matplotlib \citep{hunter2007matplotlib}, and NumPy \citep{harris2020array}.

\bibliography{reference}
\bibliographystyle{iclr2023_conference}

\appendix

\newpage

\section{Proofs}

\subsection{Proof of Theorem 1}
\label{appendix: thm1}

\begin{proof}
Note that for any pair $(\bm{s}^{(l)}, {V}^{(l)})$  with $\text{diag}( \bm{s}^{(l)})  \cdot \bm{V}^{(l)} = W^{(l)}$, we have
\begin{align}
    \|\bm{V}^{(l)}\| \longrightarrow 0, \text{ as } \bm{s}^{(l)} \longrightarrow \infty
\end{align}
for $l = 1, 2, \dots, L+1$. Then for any $\epsilon >0$,  there exists $M >0$ such that for any $(\bm{V}^{(l)}, \bm{s}^{(l)})_{l=1}^{L+1} \in U_{\bm{\theta}}$ ,and  $\|\bm{s}^{(l)}\| > M$, 
we obtain $ \|\bm{V}^{(l)}\| < \epsilon$, for $l=1, 2, \dots, L+1$. We define $U_*$ by
\begin{align}
    U_* = \{(\bm{0}, \bm{s}^{(l)})_{l=1}^{L+1} : \bm{s}^{(l)} \in \R^{d_l}, l=1,2,\dots, L+1    \}
\end{align}
Now we can choose $\bm{s}^{(l)}$ such that $\|\bm{s}^{(l)}\| > M$. Then  
\begin{align}
    \text{dist}(U_{\bm{\theta}}, U_*) \leq \sqrt{\sum_{l=1}^{L+1} \|\bm{V}^{(l)}\|^2} \leq \sqrt{L+1} \epsilon
\end{align}
Similarly, we can show that
\begin{align}
    \text{dist}(U_{\bm{\theta}'}, U_*) \leq \sqrt{\sum_{l=1}^{L+1} \|\bm{V}^{(l)}\|^2} \leq \sqrt{L+1} \epsilon
\end{align}
Therefore, 
\begin{align}
      \text{dist}(U_{\bm{\theta}}, U_{\bm{\theta}'}) \leq + \text{dist}(U_{\bm{\theta}}, U_*) + \text{dist}(U_*, U_{\bm{\theta}'}) \leq 2\sqrt{L+1} \epsilon
\end{align}
Since $\epsilon$ is arbitrary, letting $\epsilon \rightarrow 0$ gives
\begin{align}
     \text{dist}(U_{\bm{\theta}}, U_{\bm{\theta}'}) = 0.
\end{align}
\end{proof}

\subsection{Proof of Theorem 2}
\label{appendix: thm2}

\begin{proof}
Suppose that  $f^{(k, l)}$ denotes $k$-th component of $\bm{f}^{(l)} \in \R^{d_l}$. Under the proposed weight factorization in \eqref{eq: neuron_fact}, differentiating the loss function $\mathcal{L}$ with respect to $\bm{w}^{k,l}$ and $s^{(k, l)}$, respectively, yields 
\begin{align}
     s^{(k,l)}_{n+1} &=  s^{(k,l)}_{n} - \eta \frac{\partial \mathcal{L}}{\partial   s^{(k,l)}_{n}} =  s^{(k,l)}_{n} - \eta \frac{\partial \mathcal{L}}{\partial f^{(k, l)}} \cdot 
     \bm{v}^{(k,l)}_n \cdot \bm{g}^{(l-1)}, \\
      \bm{v}^{(k,l)}_{n+1} &=   \bm{v}^{(k,l)}_{n} - \eta \frac{\partial \mathcal{L}}{\partial   \bm{v}^{(k,l)}_{n}} = \bm{v}^{(k,l)}_{n} - \eta s^{(k,l)}_n \frac{\partial \mathcal{L}}{\partial f^{(k, l)}} \cdot \bm{g}^{(l-1)}.
\end{align}
Note that
\begin{align}
    \frac{\partial \mathcal{L}}{\partial \bm{w}^{(k,l)}_{n}} = \frac{\partial \mathcal{L}}{\partial f^{(k, l)}} \cdot \bm{g}^{(l-1)},
\end{align}
and the update rule of $\bm{v}^{(k,l)}$ and $ s^{(k,l)}$ can be re-written as
\begin{align}
      s^{(k,l)}_{n+1} &=  s^{(k,l)}_{n} - \eta \bm{v}^{(k,l)}_n \cdot \frac{\partial \mathcal{L}}{\partial \bm{w}^{(k,l)}_{n}}, \\
       \bm{v}^{(k,l)}_{n+1} &=  \bm{v}^{(k,l)}_{n} - \eta s^{(k,l)}_n   \frac{\partial \mathcal{L}}{\partial \bm{w}^{(k,l)}_{n}}.
\end{align}
Since $\bm{w}^{(k, l)} = s^{(k, l)} \cdot \bm{v}^{(k, l)} $, the update rule of $\bm{w}^{(k, l)}$ is given by
\begin{align}
    \bm{w}^{(k, l)}_{n+1} = \bm{w}^{(k, l)}_{n} - \eta \left(\|[s^{(k,l)}_n + \bm{v}^{(k,l)}_n\|^2_2 ]^2 \right)  \frac{\partial \mathcal{L}}{\partial \bm{w}^{(k,l)}_{n}}  + \mathcal{O}(\eta^2)
\end{align}
\end{proof}

\section{Algorithm of Random weight factorization}
\label{appendix: alg}

\begin{algorithm}[H]
\caption{Random weight factorization (RWF)}\label{alg}
\begin{algorithmic}
\State {1. Initialize a neural network $f_{\bm{\theta}}$ with $\bm{\theta} = \{\bm{W}^{(l)}, \bm{b}^{(l)}\}_{l=1}^{L+1}$ (e.g. using the Glorot scheme \citep{glorot2010understanding}).}
      \For{$l = 1,2, \dots, L$}
        \State {(a) Initialize each scale factor as $\bm{s}^{(l)}\sim\mathcal{N}(\mu, \sigma I)$.} 
        \State {(b)  Construct the factorized weight matrices as $\bm{W}^{(l)} = \text{diag}( \exp( \bm{s}^{(l)})) \cdot \bm{V}^{(l)}$.  }
      \EndFor
      \State {2. Train the network by gradient descent on the factorized parameters $\{\bm{s}^{(l)}, \bm{V}^{(l)}, \bm{b}^{(l)}\}_{l=1}^{L+1}$.}

\State {The recommended hyper-parameters are $\mu=1.0, \sigma=0.1$.} 
\end{algorithmic}
\end{algorithm}

\section{A drop-in enhancement for linear layers}
\label{appendix: flax}

\begin{lstlisting}[language=Python, caption=JAX Flax implementation \citep{flax2020github} of a conventional linear layer.]
class Dense(nn.Module):
    features: int
        
    @nn.compact
    def __call__(self, x):
        kernel = self.param('kernel', 
                            glorot_normal(), 
                            (x.shape[-1], 
                             self.features))
        bias = self.param('bias', 
                          nn.initializers.zeros, 
                          (self.features,))
        y = np.dot(x, kernel) + bias
        return y
\end{lstlisting}

\begin{lstlisting}[language=Python, caption=JAX Flax implementation \citep{flax2020github} of a linear layer with \textit{random weight factorization}.]
def factorized_glorot_normal(mean=1.0, stddev=0.1):
    def init(key, shape):
        key1, key2 = random.split(key)
        w = glorot_normal()(key1, shape) 
        s = mean + normal(stddev)(key2, (shape[-1],))
        s = np.exp(s)
        v = w / s
        return s, v
    return init

class FactorizedDense(nn.Module):
    features: int

    @nn.compact
    def __call__(self, x):
        s, v = self.param('kernel', 
                          factorized_glorot_normal(), 
                          (x.shape[-1], self.features)) 
        kernel = s * v
        bias = self.param('bias', 
                          nn.initializers.zeros, 
                          (self.features,))
        y = np.dot(x, kernel) + bias
        return y
\end{lstlisting}

\section{1D Regression Ablation Study}
\label{appendix: 1d_regression}

\begin{figure}[H]
    \centering
    \includegraphics[width=0.8\textwidth]{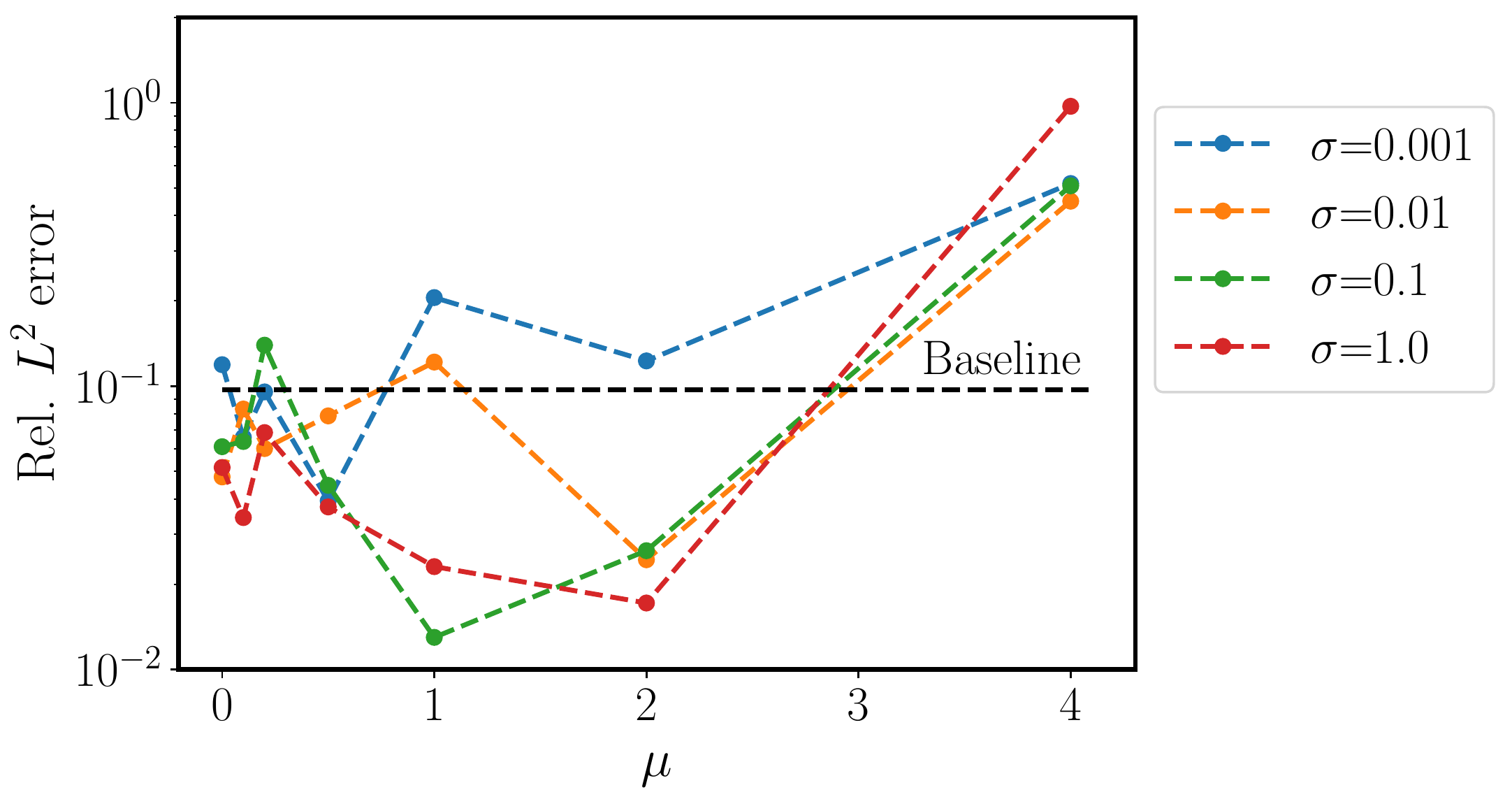}
    \caption{{\em 1D Regression:} Resulting relative $L^2$ errors of training MLPs with \textit{random weight factorization} initialized by different  $\mu$ and $\sigma$. The baseline (black dash) represents the result of training a plain MLP.}
    \label{fig: 1d_regression_sys}
\end{figure}

We perform a systematic study on the effect of $\mu$ and $\sigma$ used for initializing the scale factor $s$ in \textit{random weight factorization}. To this end, we train MLPs with \textit{random weight factorization} initialized by different  $\mu$ and $\sigma$. Each model (3 layers, 128 channels, ReLU activations) is trained via a full-batch gradient descent for $10^5$ iterations using the Adam optimizer \citep{kingma2014adam} with a starting learning rate of $10^{-3}$ followed by an exponential decay by a factor of 0.9 in every $5,000$ steps. The resulting relative $L^2$ errors are visualized in Figure  \ref{fig: 1d_regression_sys}. We observe that models initialized with a small $\mu$ achieve similar performance to our baseline, while large $\mu$ and $\sigma$ values can lead to an unstable training process and poor predictive accuracy. In particular, $\mu=1$ and $\sigma =0.1$ yield the best results, and thus we will use this as the default hyper-parameter of random weight factorization in the majority of the benchmarks presented here (see Table \ref{tab: rwf}).

\clearpage
\section{Hyper-parameters}
\label{appendix: hyper_parameters}

The following tables summarizes the hyper-parameters of the different networks architectures employed in each benchmark (Table \ref{tab: network}), their associated learning rate schedules (Table \ref{tab: lr}), and the corresponding random weight factorization settings (Table \ref{tab: rwf}).

\begin{table}[H]
    \centering
     \renewcommand{\arraystretch}{1.2}
     \resizebox{\textwidth}{!}{
    \begin{tabular}{|c|c|c|c|c|c|}
\hline
Task                                    & Case                           & Backbone                  & Depth              & width                & Activation            \\ \hline
\multirow{2}{*}{Image Regression}       & Natural                        & \multirow{2}{*}{MLP}      & \multirow{2}{*}{4} & \multirow{2}{*}{256} & \multirow{2}{*}{ReLU} \\ \cline{2-2}
                                        & Text                           &                           &                    &                      &                       \\ \hline
\multirow{2}{*}{Shape Representation}   & Dragon                         & \multirow{2}{*}{MLP}      & \multirow{2}{*}{8} & \multirow{2}{*}{256} & \multirow{2}{*}{ReLU} \\ \cline{2-2}
                                        & Armadillo                      &                           &                    &                      &                       \\ \hline

\multirow{2}{*}{Computed Tomography} & Shepp                          & \multirow{2}{*}{MLP}      & \multirow{2}{*}{5} & \multirow{2}{*}{256} & \multirow{2}{*}{ReLU} \\ \cline{2-2}
                                        & ATLAS                          &                           &                    &                      &                       \\ \hline
Inverse Rendering                       & Lego                           & MLP                       & 5                  & 256                  & ReLU                  \\ \hline
\multirow{3}{*}{Solving PDEs}           & Advection                      & MLP                       & 5                  & 256                  & \multirow{3}{*}{Tanh} \\ \cline{2-5}
                                        & \multirow{2}{*}{Navier-Stokes} & MLP                       & \multirow{2}{*}{5} & \multirow{2}{*}{128} &                       \\ \cline{3-3}
                                        &                                & Modified MLP              &                    &                      &                       \\ \hline
\multirow{3}{*}{Learning Operators}     & DR                             & \multirow{2}{*}{DeepONet} & 5                  & 64                   & Tanh                  \\ \cline{2-2} \cline{4-6} 
                                        & Darcy                          &                           & 4                  & 128                  & GELU                  \\ \cline{2-6} 
                                        & Burgers                        & Modified  DeepONet        & 5                  & 128                  & Tanh                  \\ \hline
\end{tabular}
}
    \caption{Network architectures for each benchmark.}
    \label{tab: network}
\end{table}

\begin{table}[H]
    \centering
    \renewcommand{\arraystretch}{1.2}
      \resizebox{\textwidth}{!}{
\begin{tabular}{|c|c|cccc|c|}
\hline
\multirow{2}{*}{Task}                 & \multirow{2}{*}{Case}          & \multicolumn{4}{c|}{Learning Rate  Schedule}                                                                                                               & \multirow{2}{*}{Iterations} \\ \cline{3-6}
                                      &                                & \multicolumn{1}{c|}{Step Size}             & \multicolumn{1}{c|}{Decay Steps}           & \multicolumn{1}{c|}{Decay Rate}           & Warmup Steps         &                             \\ \hline
\multirow{2}{*}{Image Regression}     & Natural                        & \multicolumn{1}{c|}{\multirow{2}{*}{$10^{-3}$}} & \multicolumn{1}{c|}{\multirow{2}{*}{-}}    & \multicolumn{1}{c|}{\multirow{2}{*}{-}}   & \multirow{2}{*}{$2 \times 10^2$} & \multirow{2}{*}{$2 \times 10^3$}       \\ \cline{2-2}
                                      & Text                           & \multicolumn{1}{c|}{}                      & \multicolumn{1}{c|}{}                      & \multicolumn{1}{c|}{}                     &                      &                             \\ \hline
\multirow{2}{*}{Shape Representation} & Dragon                         & \multicolumn{1}{c|}{\multirow{2}{*}{$5 \times 10^{-4}$}} & \multicolumn{1}{c|}{\multirow{2}{*}{$5 \times 10^3$}}    & \multicolumn{1}{c|}{\multirow{2}{*}{0.1}}   & \multirow{2}{*}{$10^3$}   & \multirow{2}{*}{$10^4$}       \\ \cline{2-2}
                                      & Armadillo                      & \multicolumn{1}{c|}{}                      & \multicolumn{1}{c|}{}                      & \multicolumn{1}{c|}{}                     &                      &                             \\ \hline
\multirow{2}{*}{Computed Tomography}  & Shepp                          & \multicolumn{1}{c|}{\multirow{2}{*}{$10^{-3}$}} & \multicolumn{1}{c|}{\multirow{2}{*}{-}}    & \multicolumn{1}{c|}{\multirow{2}{*}{-}}   & \multirow{2}{*}{$2 \times 10^2$} & \multirow{2}{*}{$2 \times 10^3$}       \\ \cline{2-2}
                                      & ATLAS                          & \multicolumn{1}{c|}{}                      & \multicolumn{1}{c|}{}                      & \multicolumn{1}{c|}{}                     &                      &                             \\ \hline
Inverse Rendering                     & Lego                           & \multicolumn{1}{c|}{$10^{-3}$}                  & \multicolumn{1}{c|}{$10^3$}                  & \multicolumn{1}{c|}{0.9}                  & $5 \times 10^3$                 & $5 \times 10^4$                       \\ \hline
\multirow{3}{*}{Solving PDEs}         & Advection                      & \multicolumn{1}{c|}{\multirow{3}{*}{$10^{-3}$}} & \multicolumn{1}{c|}{$5 \times 10^3$}                  & \multicolumn{1}{c|}{0.9}                  & \multirow{3}{*}{-}   & $2 \times 10^5$                     \\ \cline{2-2} \cline{4-5} \cline{7-7} 
                                      & \multirow{2}{*}{Navier-Stokes} & \multicolumn{1}{c|}{}                      & \multicolumn{1}{c|}{\multirow{2}{*}{$2 \times 10^3$}} & \multicolumn{1}{c|}{\multirow{2}{*}{0.9}} &                      & \multirow{2}{*}{$1.2 \times 10^5$}     \\
                                      &                                & \multicolumn{1}{c|}{}                      & \multicolumn{1}{c|}{}                      & \multicolumn{1}{c|}{}                     &                      &                             \\ \hline
\multirow{3}{*}{Learning Operators}   & DR                             & \multicolumn{1}{c|}{\multirow{3}{*}{$10^{-3}$}} & \multicolumn{1}{c|}{\multirow{2}{*}{$ 10^3$}} & \multicolumn{1}{c|}{\multirow{2}{*}{0.9}} & \multirow{3}{*}{-}   & \multirow{2}{*}{$5 \times 10^4$}      \\ \cline{2-2}
                                      & Darcy                          & \multicolumn{1}{c|}{}                      & \multicolumn{1}{c|}{}                      & \multicolumn{1}{c|}{}                     &                      &                             \\ \cline{2-2} \cline{4-5} \cline{7-7} 
                                      & Burgers                        & \multicolumn{1}{c|}{}                      & \multicolumn{1}{c|}{$2 \times 10^3$}                  & \multicolumn{1}{c|}{0.9}                  &                      & $ 10^5$                      \\ \hline
\end{tabular}
}
    \caption{Learning Rate Schedules for each benchmark.}
    \label{tab: lr}
\end{table}

\begin{table}[H]
    \centering
        \renewcommand{\arraystretch}{1.2}
    \begin{tabular}{|c|c|c|}
\hline
Task                                  & Case          & Initialization of RWF \\ \hline
\multirow{2}{*}{Image Regression}     & Natural       & $\bm{s} \sim \mathcal{N}(2, 0.01) $                         \\ \cline{2-3} 
                                      & Text          &$\bm{s} \sim \mathcal{N}(1, 0.1) $                   \\ \hline
\multirow{2}{*}{Shape Representation} & Dragon        & \multirow{2}{*}{$\bm{s} \sim \mathcal{N}(1, 0.1) $}            \\ \cline{2-2}
                                      & Armadillo     &                                       \\ \hline
\multirow{2}{*}{Computed Tomography}  & Shepp         & \multirow{2}{*}{$\bm{s} \sim \mathcal{N}(1, 0.1) $}            \\ \cline{2-2}
                                      & ATLAS         &                                       \\ \hline
Inverse Rendering                     & Lego          & $\bm{s} \sim \mathcal{N}(1, 0.1) $                       \\ \hline
\multirow{2}{*}{Solving PDEs}         & Advection     & $\bm{s} \sim \mathcal{N}(1, 0.1) $                       \\ \cline{2-3} 
                                      & Navier-Stokes & $\bm{s} \sim \mathcal{N}(0.5, 0.1) $                      \\ \hline
\multirow{3}{*}{Learning Operators}   & DR            & \multirow{3}{*}{$\bm{s} \sim \mathcal{N}(1, 0.1) $}            \\ \cline{2-2}
                                      & Darcy         &                                       \\ \cline{2-2}
                                      & Burgers       &                                       \\ \hline
\end{tabular}
    \caption{Distribution used for initializing the scale factor in \textit{random weight factorization}.}
    \label{tab: rwf}
\end{table}

\section{Computational Costs}
\label{appendix: computational_cost}

Table \ref{tab: computational_cost} presents the computational cost in terms of training iterations per second for the networks networks employed in each benchmark. All timings are reported on a single NVIDIA RTX A6000 GPU. 

\begin{table}[H]
    \centering
        \renewcommand{\arraystretch}{1.2}
    \begin{tabular}{|c|c|c|c|c|c|}
\hline
Task                                    & Case          & Plain  & AA     & WN     & RWF    \\ \hline
\multirow{2}{*}{Image Regression}       & Natural       & 116.36 & 111.33 & 114.09 & 113.80 \\ \cline{2-6} 
                                        & Text          & 116.86 & 112.04 & 113.77 & 113.99 \\ \hline
\multirow{1}{*}{Shape Representation}   & Dragon        &  12.79 &  12.79 &  12.76  &  12.73      \\ \cline{2-6} 
                                        & Armadillo     &  13.14 & 13.06  & 12.81  & 12.94  \\  \hline
\multirow{2}{*}{Computed Tomography} & Shepp            & 30.02  & 29.21  & 29.83  & 29.88  \\ \cline{2-6} 
                                        & ATLAS         & 29.47  & 28.74  & 29.36  & 29.32  \\ \hline
Inverse Rendering                       & Lego          & 30.59  & 29.41  & 30.68  & 30.75  \\ \hline
Solving PDEs                            & Advection     & 853.82 & 757.19 & 855.50 & 789.41 \\ \hline
                                        & Navier-Stokes & 164.02 & 152.70 & 160.51 & 160.45 \\ \hline
\multirow{3}{*}{Learning Operators}     & DR            & 469.09 & 450.44 & 470.34 & 469.00 \\ \cline{2-6} 
                                        & Darcy         & 63.10  & 61.21  & 62.92  & 62.91  \\ \cline{2-6} 
                                        & Burgers       & 27.86  & 26.25  & 27.29  & 27.10  \\ \hline
\end{tabular}
    \caption{Computational cost (training iterations per second) for each benchmark. We can see that  the computational overhead of \textit{random weight factorization} is marginal. }
    \label{tab: computational_cost}
\end{table}

\clearpage

\section{2D Image regression}
\label{appendix: 2d_image}

As mentioned in Section \ref{sec: 2d_img},  we  use two image data-sets: \textit{Natural} and \textit{Natural}. All the test images have a $512 \times 512$ resolution while the training data has a $256 \times 256$ resolution. For each data-set, we compare the performance of MLPs with different parameterizations (see \ref{sec: A.A&W.N}) and the following input embeddings:

\textbf{No mapping:}  MLP with no input feature mapping.

\textbf{Positional encoding  \citep{tancik2020fourier}:} $\gamma(\bm{x})=\left[\ldots, \cos \left(2 \pi \sigma^{j / m} \bm{x}\right), \sin \left(2 \pi \sigma^{j / m} \bm{x}\right), \ldots\right]^{\mathrm{T}}$ for $j=0, 1, \dots, m-1$. where the frequencies are log-linear spaced and the scale $\sigma >0$ is a user-specified hyper-parameter. 

\textbf{Gaussian \citep{tancik2020fourier}:} $\gamma(\bm{x})=[\cos (2 \pi \mathbf{B} \bm{x}), \sin (2 \pi \mathbf{B} \bm{x})]^{\mathrm{T}}$, where $\mathbf{B} \in \R^{m \times d}$ is sampled from a Gaussian distribution $\mathcal{N}(0, \sigma^2)$. The scale $\sigma >0$ is a user-specified hyper-parameter.

\begin{figure}[H]
     \centering
     \begin{subfigure}[b]{0.9\textwidth}
         \centering
         \includegraphics[width=\textwidth]{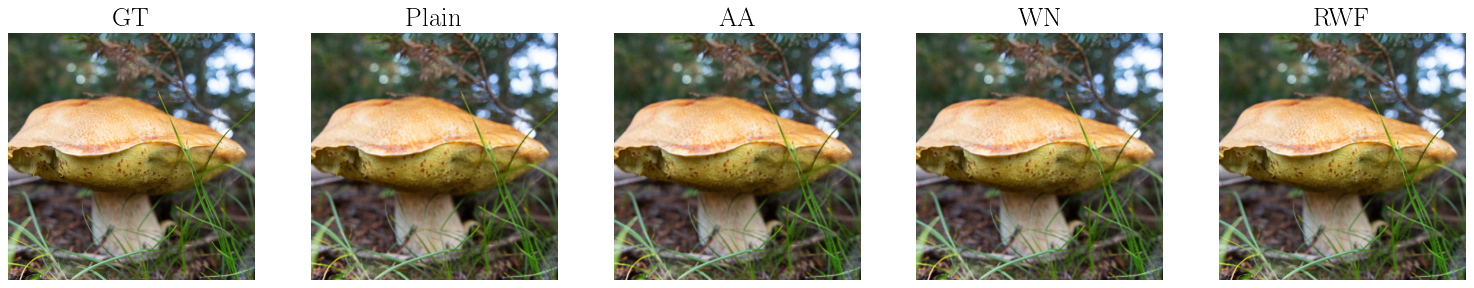}
     \end{subfigure}
     \hfill
      \begin{subfigure}[b]{0.9\textwidth}
         \centering
         \includegraphics[width=\textwidth]{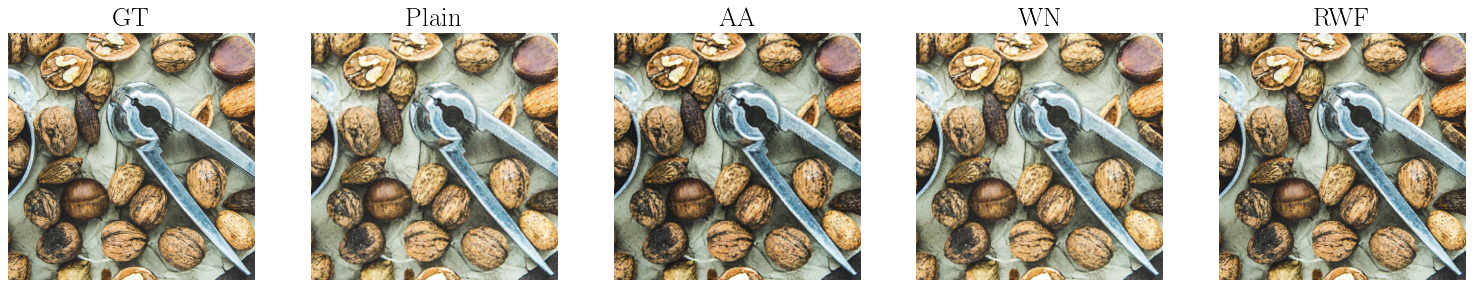}
     \end{subfigure}
     \hfill
        \caption{{\em 2D Image Regression:} Predicted \textit{Natural} images of trained MLPs with Gaussian Fourier features and with different weight parameterizations.}
        \label{fig: 2d_image_natural}
\end{figure}

\begin{figure}[H]
     \centering
     \begin{subfigure}[b]{0.9\textwidth}
         \centering
         \includegraphics[width=\textwidth]{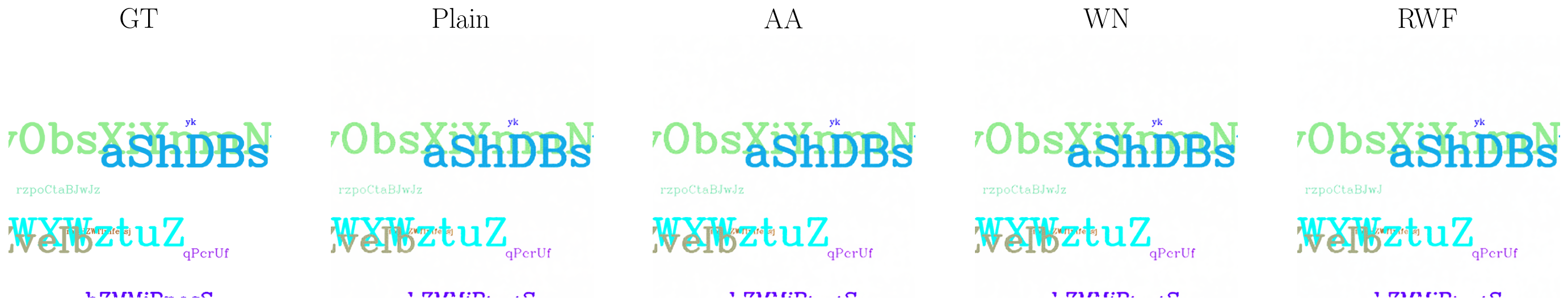}
     \end{subfigure}
      \begin{subfigure}[b]{0.9\textwidth}
         \centering
         \includegraphics[width=\textwidth]{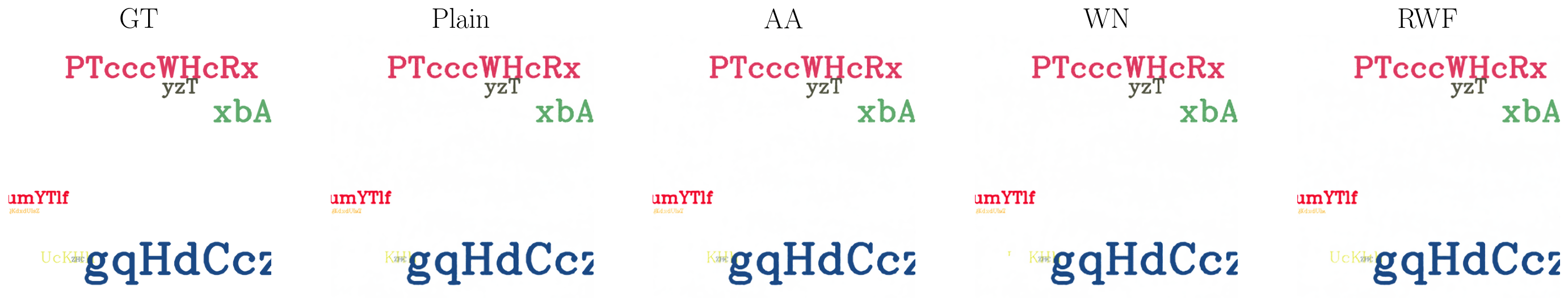}
     \end{subfigure}
     \hfill
          \caption{{\em 2D Image Regression:} Predicted \textit{Text} images of trained MLPs with Gaussian Fourier features and with different weight parameterizations.}
        \label{fig: 2d_image_text}
\end{figure}

Each model (4 layers, 256 channels, ReLU activations) is trained via a full-batch gradient descent for 2,000 iterations using the Adam optimizer \citep{kingma2014adam} with default settings and 200 warm-up steps. Particularly, the mapping scales of positional encoding and Gaussian Fourier features are the same as in \citep{tancik2020fourier}. For random weight factorization, we initialize the scale factor $\bm{s} \sim \mathcal{N}(2, 0.01)$ and $\bm{s} \sim \mathcal{N}(1, 0.01)$ for the \textit{Natural} and \textit{Text}  data-set, respectively.
The resulting test PSNR is reported in Table \ref{tab: 2d_image_regression_natural} and Table \ref{tab: 2d_image_regression_text}. We can see that the weight factorization with Gaussian Fourier features achieves the best PSNR among all the cases. Some visualizations are shown in Figure \ref{fig: 2d_image_natural} and Figure \ref{fig: 2d_image_text}.

\begin{table}[H]
    \centering
    \renewcommand{\arraystretch}{1.2}
    \begin{tabular}{c|c|c|c|c}
    \hline
        \textit{Natural} data-set         & Plain & AA  & WN & RWF (ours) \\
       \hline
       \hline
      No mapping   & $17.94 \pm 2.38$  & $18.29 \pm 2.44$ & $18.28 \pm 2.44$ &  $\mathbf{19.11 \pm 2.50}$\\
       Positional Encoding & $27.04 \pm 3.89 $  & $26.73 \pm 3.67$ & $26.99 \pm 3.83$ & $\mathbf{27.46 \pm 3.82}$\\
      Gaussian & $27.35 \pm 4.05$  & $27.36 \pm 3.96$ & $27.36 \pm 4.03$ & $\mathbf{28.08 \pm 4.34}$ \\
    \end{tabular}
    \caption{{\em 2D Image Regression:} Mean and standard deviation of PSNR obtained by training MLPs with different input mappings and weight parameterizations for the \textit{Natural} data-set.}
    \label{tab: 2d_image_regression_natural}
\end{table}

\begin{table}[H]
    \centering
    \renewcommand{\arraystretch}{1.2}
    \begin{tabular}{c|c|c|c|c}
    \hline
         \textit{Text} data-set         & Plain & AA  & WN & RWF (ours) \\
       \hline
       \hline
      No mapping   & $18.42 \pm 2.42$  & $18.43 \pm 2.34$ & $\mathbf{18.46 \pm 2.31}$ &  $17.85 \pm 2.36$\\
       Positional encoding & $31.33 \pm 2.71 $  & $\mathbf{31.73 \pm 2.29}$ & $31.43 \pm 2.57$ & $30.49 \pm 2.45$\\
      Gaussian & $32.09 \pm 1.80$  & $32.29 \pm 1.99$ & $32.25 \pm 1.74$ & $\mathbf{33.13 \pm 2.03}$ \\
    \end{tabular}
   \caption{{\em 2D Image Regression:} Mean and standard deviation of PSNR obtained by training MLPs with different input mappings and weight parameterizations for the \textit{Text} data-set.}
    \label{tab: 2d_image_regression_text}
\end{table}

\begin{figure}[H]
     \centering
     \begin{subfigure}[b]{0.4\textwidth}
         \centering
         \includegraphics[width=\textwidth]{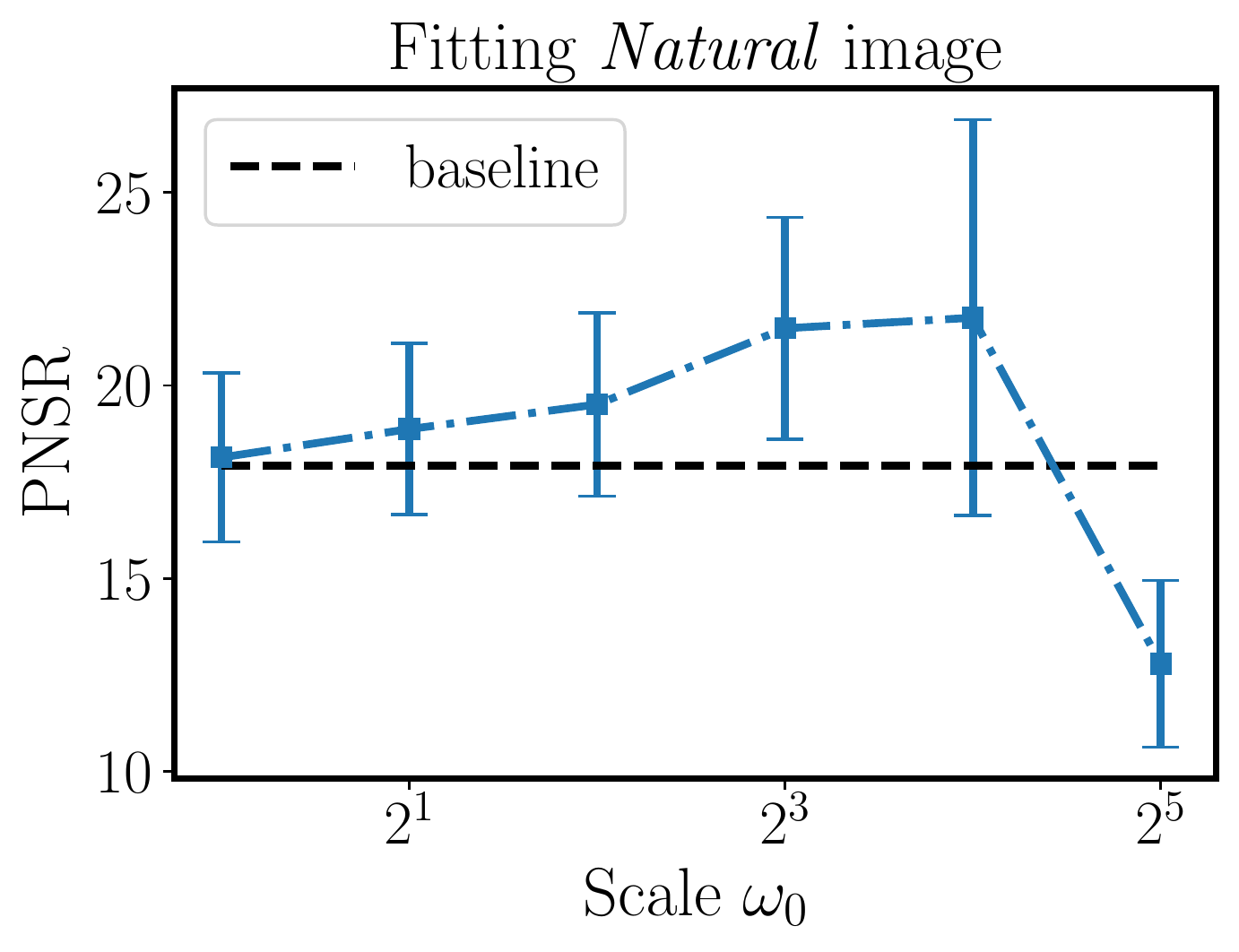}
     \end{subfigure}
     \begin{subfigure}[b]{0.4\textwidth}
         \centering
         \includegraphics[width=\textwidth]{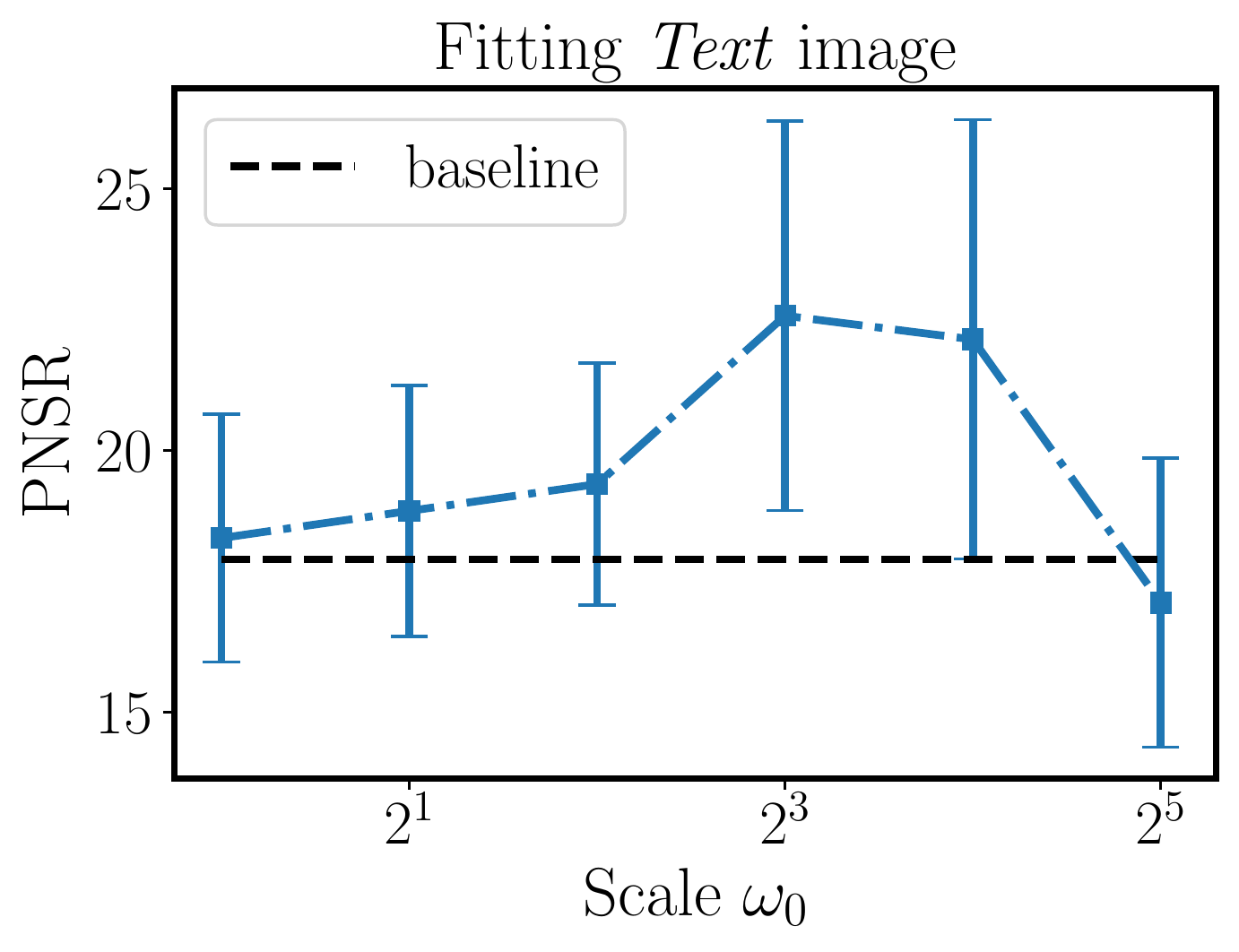}
     \end{subfigure}
        \caption{{\em 2D Image Regression:} PSNR of training SIREN networks with different scale factor for the \textit{Natural} and \textit{Text}  data-set, respectively. Error bars are plotted over different images in the data-set. The baseline (black dash) represents the result of training a plain MLP with no input mapping.}
        \label{fig: 2d_image_siren}
\end{figure}

\paragraph{Comparison with SIREN \citep{sitzmann2020implicit}:} We find that SIREN also factorizes the weight matrix of every hidden layer as $W = \omega_0 \times \widehat{W}$ with some scale factor $\omega_0$. It is indeed a special case of our approach. To examine its performance, we vary the scale factor $w_0$ and train SIREN networks under the same hyper-parameter setting, and present the test error over the \textit{Natural} and \textit{Text} data-set in Figure \ref{fig: 2d_image_siren}. It can be observed that the scale factor $w_0$ plays a fundamental role in the SIREN performance. If we take $w_0 = 1$, then SIREN just performs similarly to our baseline (Plain MLP with no input mapping). Therefore, we may argue that the success of SIREN can be attributed to that simple weight factorization instead of the sine activations with the associated initialization scheme $W \sim \mathcal{U}(-\sqrt{6 / d}, \sqrt{6 / d})$. Nevertheless, the best PSNR that SIREN achieves is still significantly lower than the proposed random weight factorization with positional encodings or Gaussian Fourier features.

\section{2D Computed Tomography}
\label{appendix: CT}

For this task, we use two data-sets: procedurally-generated 
Shepp-Logan phantoms \citep{shepp1974fourier} and 2D brain images from the ATLAS data-set  \citep{liew2018large}. Each data-set consist of 20  images of $512 \times 512$ resolution. To generate the training data, we compute 20 and 40 synthetic integral projections at evenly-spaced angles for every image of \textit{Shepp} and \textit{ATLAS} data-set, respectively.

Similar to the previous tasks, we compare the performance of MLPs with different input mappings and weight parameterizations. In experiments,
we take a scale factor $\sigma=3$ for both positional encoding and Gaussian Fourier features and initialize $\bm{s} \sim \mathcal{N}(1, 0.1) $ for  using random weight factorization.
Each model (5 layers, 256 channels, ReLU activations) is trained via a full-batch gradient descent for 2,000 iteration using the the Adam optimizer \citep{kingma2014adam} \citep{kingma2014adam} with default settings and 200 warm-up steps. 

\begin{figure}[H]
     \centering
     \begin{subfigure}[b]{0.9\textwidth}
         \centering
         \includegraphics[width=\textwidth]{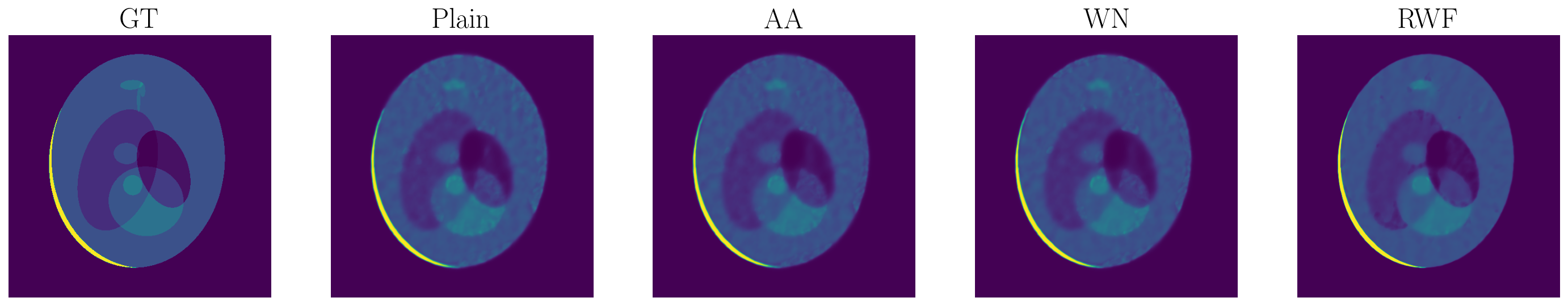}
     \end{subfigure}
      \begin{subfigure}[b]{0.9\textwidth}
         \centering
         \includegraphics[width=\textwidth]{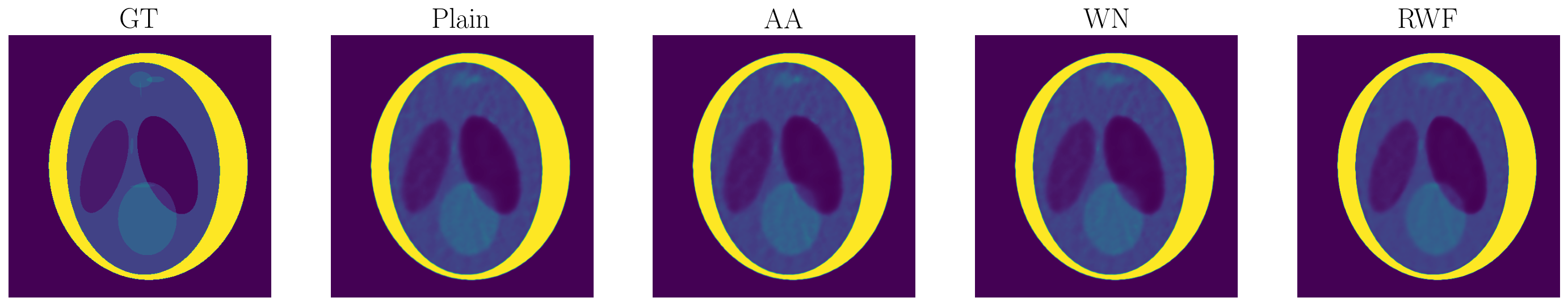}
     \end{subfigure}
     \hfill
       \caption{{\em 2D  Computed Tomography:} Predictions of trained MLPs with Gaussian Fourier features and with different weight parameterizations for the \textit{Shepp} data-set.}
        \label{fig: 2d_ct_shepp}
\end{figure}

\begin{figure}[H]
     \centering
     \begin{subfigure}[b]{0.9\textwidth}
         \centering
         \includegraphics[width=\textwidth]{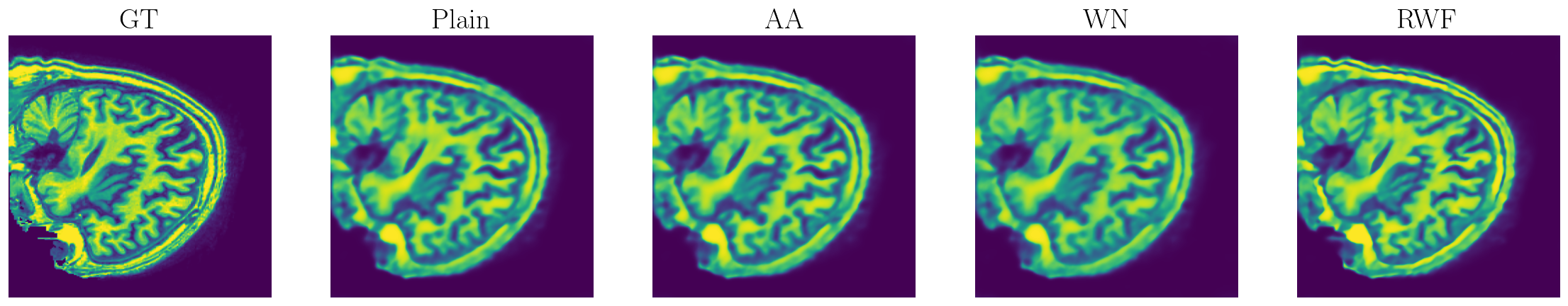}
     \end{subfigure}
      \begin{subfigure}[b]{0.9\textwidth}
         \centering
         \includegraphics[width=\textwidth]{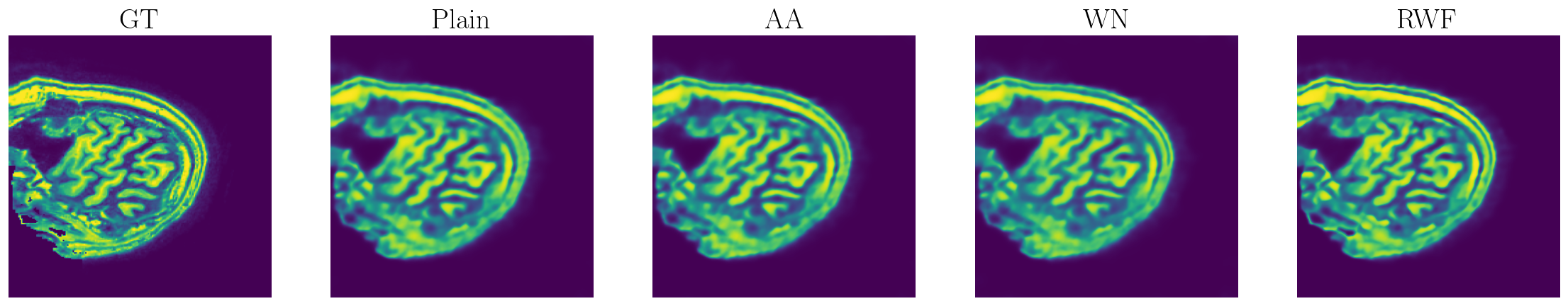}
     \end{subfigure}
     \hfill
        \caption{{\em 2D  Computed Tomography:} Predictions of trained MLPs with Gaussian Fourier features and with different weight parameterizations for the \textit{ATLAS} data-set.}
        \label{fig: 2d_ct_atlas}
\end{figure}

Table \ref{tab: 2d_ct}  summarizes the test PSRN over \textit{Shepp} and \textit{ATLAS} data-set, respectively. For different input mappings, random weight factorization yields the best PSNR, consistently outperforming other parameterizations. Besides, we plot some model predictions corresponding to Gaussian input mapping in Figure \ref{fig: 2d_ct_shepp} and \ref{fig: 2d_ct_atlas}.

\begin{table}[H]
    \centering
    \renewcommand{\arraystretch}{1.2}
    \begin{tabular}{c|c|c|c|c}
    \hline
        \textit{Shepp} data-set         & Plain & AA  & WN & RWF (ours) \\
       \hline
       \hline
      No mapping   & $22.78 \pm 1.35$  & $23.44 \pm 1.27$ & $23.37 \pm 0.84$ &  $\mathbf{24.39 \pm 1.56}$\\
       Positional encoding & $29.56 \pm 1.73 $  & $29.71 \pm 1.80$ & $29.87 \pm 1.86$ & $\mathbf{32.43 \pm 1.51}$\\
      Gaussian & $30.08 \pm 1.63$  & $30.44 \pm 1.69$ & $30.55 \pm 1.72$ & $\mathbf{33.70 \pm 1.29}$ \\
    \end{tabular}
    
    \begin{tabular}{c|c|c|c|c}
    \hline
         \textit{ATLAS} data-set     & Plain & AA  & WN & RWF (ours) \\
       \hline
       \hline
      No mapping   & $15.87 \pm 0.66$  & $16.07 \pm 0.68$ & $16.17 \pm 0.63$ &  $\mathbf{16.49 \pm 0.61}$\\
       Positional encoding & $21.44 \pm 0.94 $  & $21.54 \pm 0.92$ & $21.70 \pm 0.63$ & $\mathbf{23.34 \pm 0.92}$\\
      Gaussian & $22.02 \pm  0.93$  & $22.02 \pm 1.05$ & $22.10 \pm 1.04$ & $\mathbf{23.61 \pm 0.83}$ \\
    \end{tabular}
    
    \caption{{\em 2D  Computed Tomography:} Mean and standard deviation of PSNR obtained by training MLPs with different input mappings and weight parameterizations for the \textit{Shepp} and \textit{ATLAS} data-set, respectively.}
    \label{tab: 2d_ct}
\end{table}

\begin{figure}[H]
     \centering
     \begin{subfigure}[b]{0.4\textwidth}
         \centering
         \includegraphics[width=\textwidth]{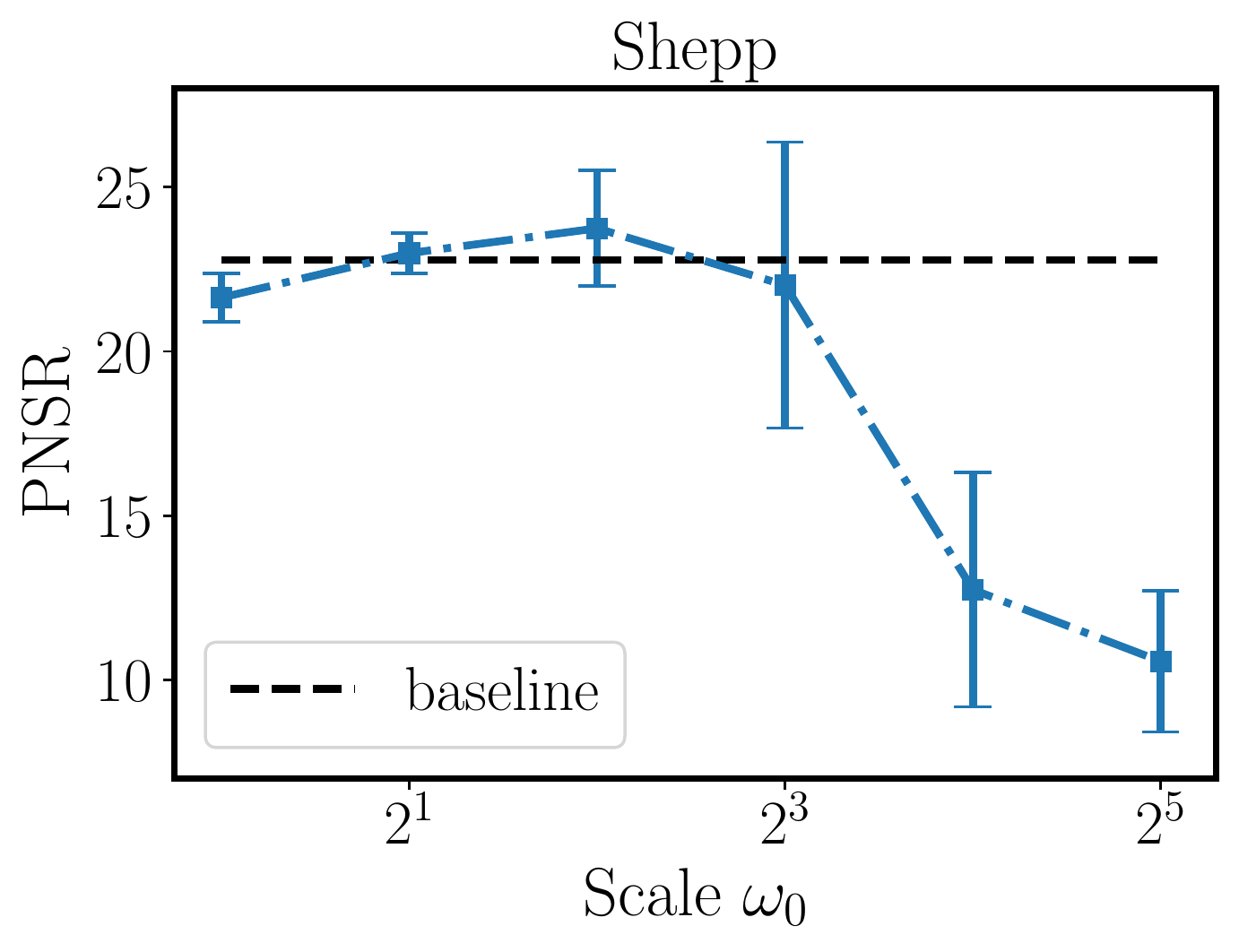}
     \end{subfigure}
     \begin{subfigure}[b]{0.4\textwidth}
         \centering
         \includegraphics[width=\textwidth]{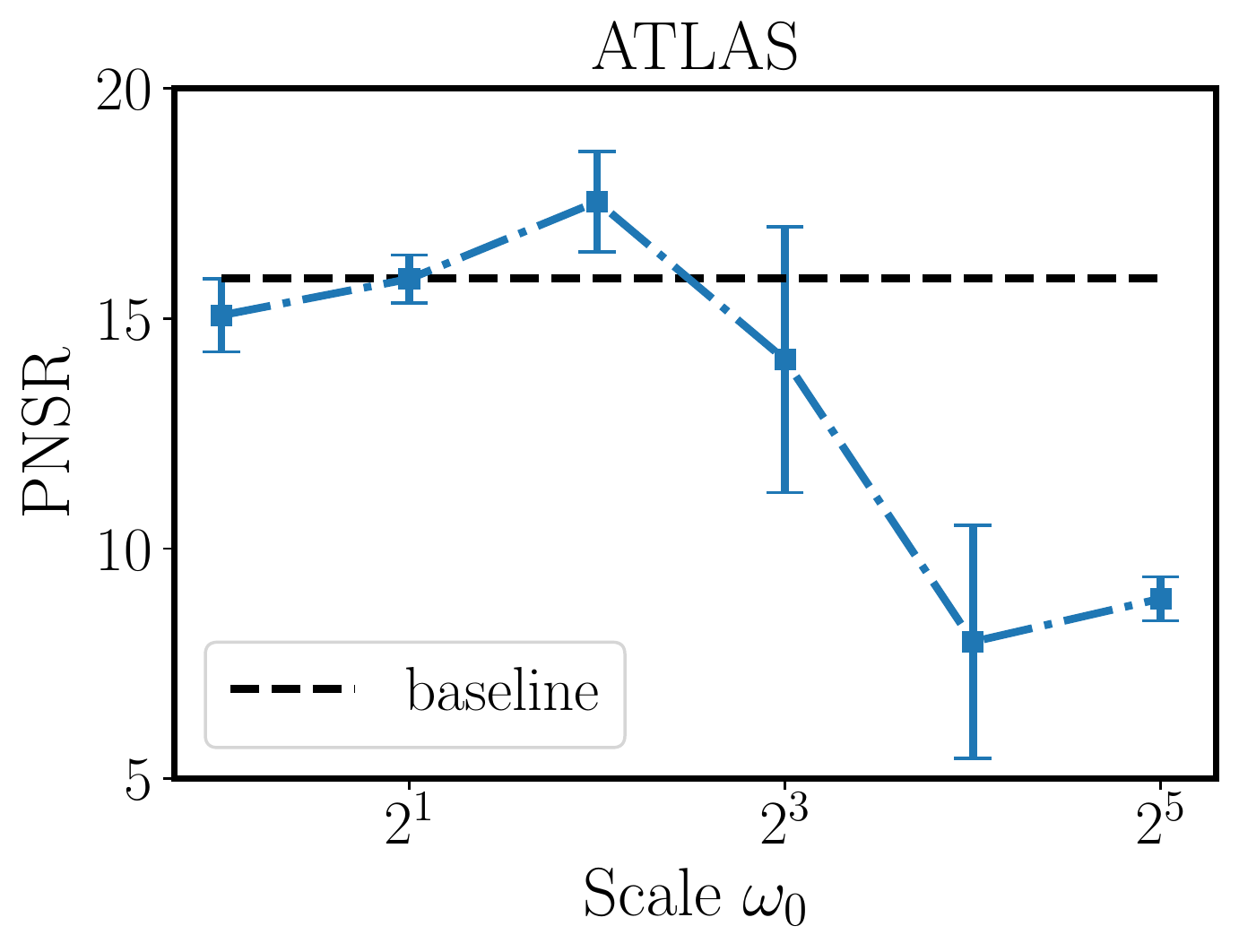}
     \end{subfigure}
        \caption{{\em 2D Computed Tomography:} PSNR of training SIREN networks with different scale factor for the \textit{Shepp} and \textit{ATLAS}  data-set, respectively. Error bars are plotted over different images in the data-set. 
        The baseline (black dash) represents the result of training a plain MLP with no input mapping.}
        \label{fig: 2d_ct_siren}
\end{figure}

\paragraph{Comparison with SIREN \citep{sitzmann2020implicit}:} We also test the performance of SIREN for this example. Specifically,  we train SIREN network with different scale factor $\omega_0$ under the same hyper-parameter settings. As shown in Figure \ref{fig: 2d_ct_siren}, SIREN using the optimal scale factor is just slightly better than our baseline (plain MLP with no input mapping), but still worse than using positional encodings or random Fourier features.

\section{3D Shape Representation}
\label{appendix: 3d_shape}

For this example, we consider two complex triangle meshes  \textit{Dragon} and \textit{Armadillo}, both of which contain hundreds of thousands of vertices.
In our experiments, every mesh is rescaled to fit inside the unit cube $[0,1]^3$ such that the centroid of the mesh is $(0.5, 0.5, 0.5)$.

We represent each shape by MLPs with different input mappings and weight parameterizations.  For models with input mappings, we use the same hyper-parameters as in \citep{tancik2020fourier}. For models using \textit{random weight factorization}, we initialize the scale factor using the recommended settings $s \sim \mathcal{N}(1, 0.1)$. All networks are trained by minimizing a cross-entropy loss to match the corresponding classification labels (0 for points outside the mesh, 1 for points inside). 

We train each model (8 layers, 128 channels, ReLU activations) via a mini-batch gradient descent for $10^4$ iterations using the Adam optimizer \citep{kingma2014adam} with a start learning rate $5 \times10^{-4}$ and an exponential decay by a factor of $0.1$ for every $5,000$ steps. The batch size we use is 8192.  To emphasize the learning of fine surface details, we calculate the test error on a set close to the mesh surface, which is generated by randomly choosing mesh vertices that have been perturbed by a random Gaussian vector with a standard deviation of 0.01.

\begin{figure}[H]
     \centering
     \begin{subfigure}[b]{0.9\textwidth}
         \centering
         \includegraphics[width=\textwidth]{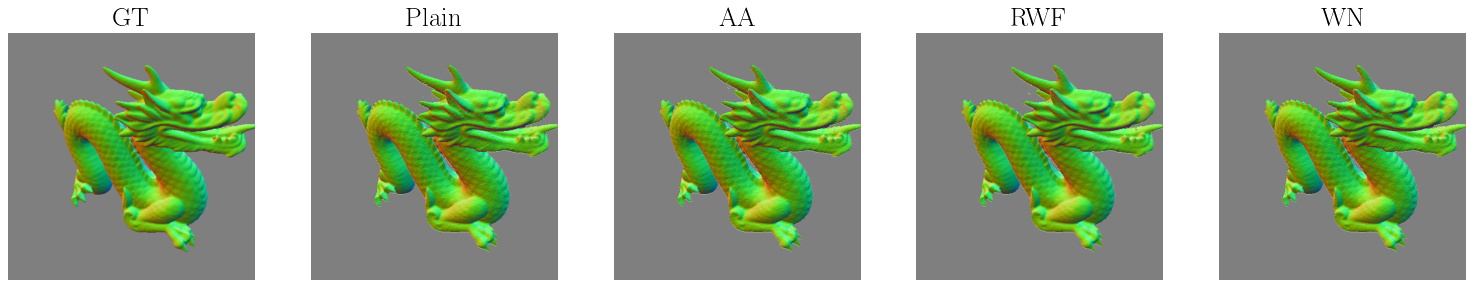}
     \end{subfigure}
      \begin{subfigure}[b]{0.9\textwidth}
         \centering
         \includegraphics[width=\textwidth]{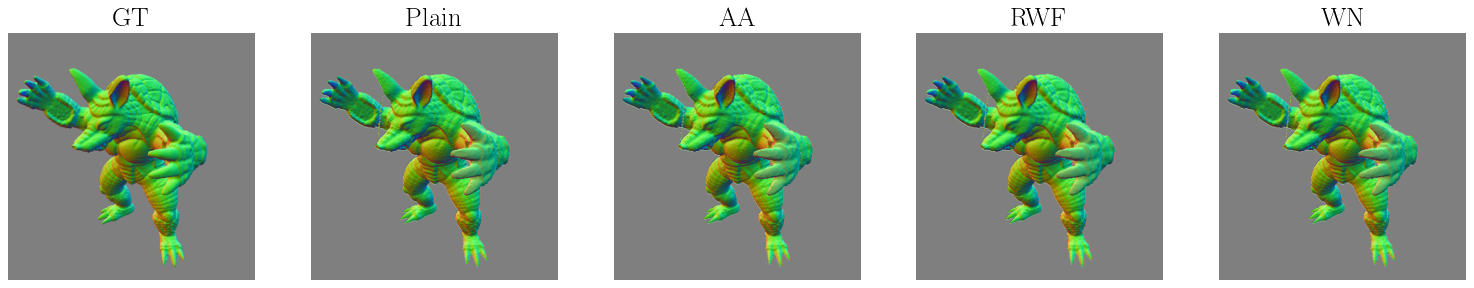}
     \end{subfigure}
     \hfill
        \caption{{\em 3D Shape Representation:} Rendered shape representations obtained by training MLPs with Gaussian Fourier features and with different weight parameterizations.}
        \label{fig: 3d_shape}
\end{figure}

The resulting IoU scores of each model is reported in Table \ref{tab: 3d_shape}. One can observe consistent improvements of RWF across different input mappings and data-sets, outperforming the other parameterizations. Moreover, the learned shape representations are depicted in Figure \ref{fig: 3d_shape}.

\begin{table}[H]
    \centering
    \renewcommand{\arraystretch}{1.2}
   
   \begin{tabular}{c|c|c|c|c}
    \hline
        \textit{Dragon} data-set         & Plain & AA  & WN & RWF (ours) \\
       \hline
       \hline
      No mapping   & 0.894  & 0.894 & 0.891 &  \textbf{0.924} \\
       Positional encoding & 0.967  & 0.968 & 0.970 & \textbf{0.977}\\
      Gaussian &  0.980  & 0.981 & 0.981 & \textbf{0.984} \\
    \end{tabular}    
    
     \begin{tabular}{c|c|c|c|c}
    \hline
        \textit{Armadillo} data-set         & Plain & AA  & WN & RWF (ours) \\
       \hline
       \hline
      No mapping   & 0.842  & 0.846 & 0.845 &  \textbf{0.901} \\
       Positional encoding & 0.965  & 0.967 & 0.967 & \textbf{0.972}\\
      Gaussian &  0.978  & 0.976 & 0.975 & \textbf{0.982} \\
    \end{tabular}

    \caption{{\em 3D Shape Representation:} IoU of training MLPs with different input mappings and weight parameterizations for the \textit{Dragon} and  \textit{Armadillo} data-sets.}
    \label{tab: 3d_shape}
\end{table}

\section{3D inverse rendering for view synthesis} 
\label{appendix: nerf}

For this task, we use the NeRF \textit{Lego} data-set of 120 images downsampled to $400 \times 400$ pixel resolution. The data-set is split into 100 training images, 7 validation images, and 13 test images. In our experiments, we only use Gaussian Fourier features  of a scale $\sigma=10$,  as it has been empirically validated to be the best input mapping in the previous tasks.

\begin{figure}[H]
    \centering
    \includegraphics[width=0.4\textwidth]{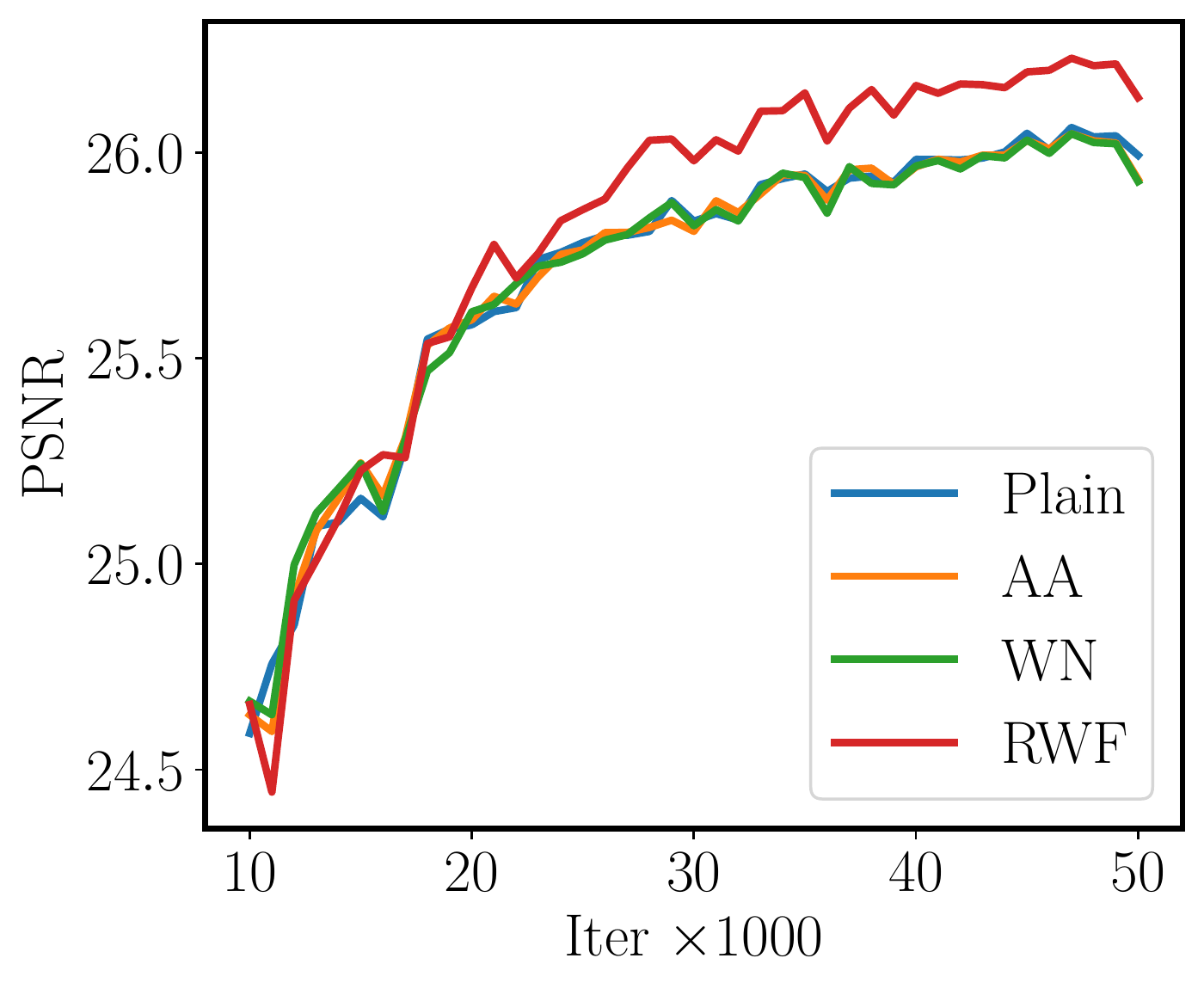}
    \caption{{\em 3D Inverse Rendering:} Test PSNR of MLPs with different weight parameterizations during training.}
    \label{fig: nerf_psnr}
\end{figure}

\begin{figure}[H]
     \centering
     \begin{subfigure}[b]{0.9\textwidth}
         \centering
         \includegraphics[width=\textwidth]{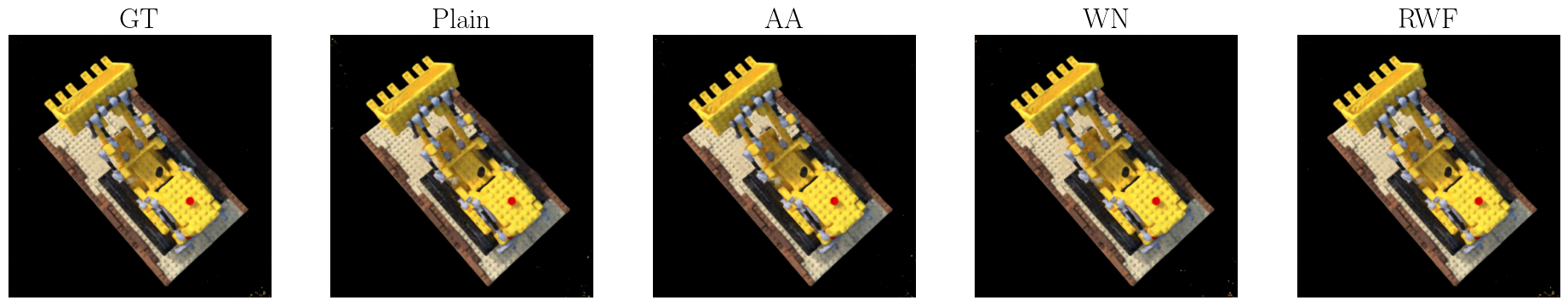}
     \end{subfigure}
      \begin{subfigure}[b]{0.9\textwidth}
         \centering
         \includegraphics[width=\textwidth]{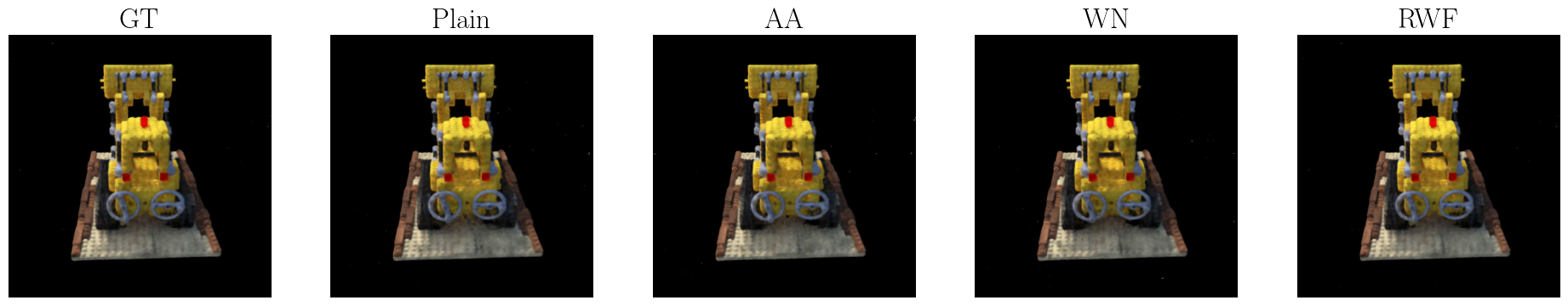}
     \end{subfigure}
     \begin{subfigure}[b]{0.9\textwidth}
         \centering
         \includegraphics[width=\textwidth]{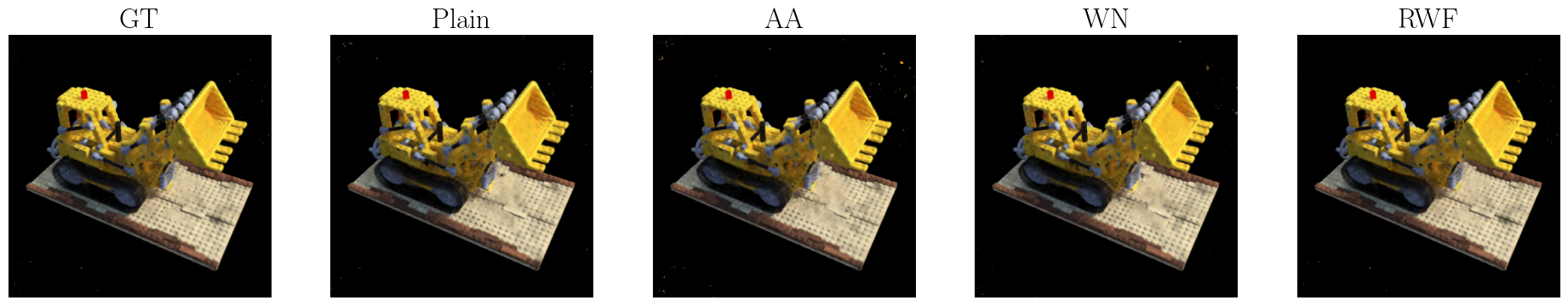}
     \end{subfigure}
        \caption{{\em 3D Inverse Rendering:}  Rendered views  of trained MLPs with different weight parameterizations.}
        \label{fig: nerf}
\end{figure}

We train MLPs (5 layers, 256 channels, ReLU activations) with different parameterizations for $5 \times 10^4$ iterations using the Adam optimizer \citep{kingma2014adam} with a start learning rate of $10^{-3}$ and a warmup exponential decay by a factor of $0.9$ for every $1,000$ steps. The batch size  is 2048. In Figure \ref{fig: nerf_psnr}, we visualize the test PSNR of each model during training. In comparison with other three parameterizations,  the MLP with RWF achieves the best PSRN. Some visualizations are provided in Figure \ref{fig: nerf}

\section{Solving PDEs}
\label{appendix: pinns}

In this section, we present the implementation details of PINNs for solving advection and Naiver-Stokes equation, respectively.

\subsection{Advection equation}

Recall 
\begin{align}
    \frac{\partial u}{\partial t}+ c \frac{\partial u}{\partial x} &=0, \quad x \in (0, 2 \pi), t \in[0, 1] \\
    u(x, 0) &=g(x), \quad x \in (0, 2 \pi)
\end{align}
with periodic boundary conditions and $c=50$.  

We represent the unknown solution $u$ by an MLP $u_{\bm{\theta}}$. In particular, we impose the exact periodic boundary conditions by constructing a special Fourier feature embedding of the form \citep{dong2021method}
\begin{align}
    \gamma(x, t) = [\cos(x), \sin(x), t]^\mathrm{T}.
\end{align}
The network can be trained by minimizing the composite loss below
\begin{align}
    \mathcal{L}(\bm{\theta}) = \lambda_{ic} \mathcal{L}_{ic}(\bm{\theta}) + \lambda_{r} \mathcal{L}_{r}(\bm{\theta})
\end{align}
where
\begin{align}
    \mathcal{L}_{ic}(\bm{\theta}) &= \frac{1}{N_{ic}} \sum_{i=1}^{N_{ic}} \left| u_{\bm{\theta}}(0, x_{ic}^i) - g(x_{ic}^i) \right|^2, \\
    \mathcal{L}_r(\bm{\theta}) &= \frac{1}{N_r} \sum_{i=1}^{N_r} \left| \frac{\partial u_{\bm{\theta}}}{\partial t}(t_r^i, x_r^i) + c \frac{\partial u_{\bm{\theta}}}{\partial x}(t_r^i, x_r^i) \right|^2. 
\end{align}
Here we set $N_{ic} = 128$ and $N_r = 1024$, and  $\{x_{ic}\}_{i=1}^{N_{ic}}, \{(x_r, t_r)\}_{i=1}^{N_r}$ are randomly sampled from the computational domain, respectively, at each iteration of gradient descent. In addition, we take $\lambda_{ic} = 100, \lambda_r = 1$ for better enforcing the initial condition. It is worth pointing out that all the network derivatives are computed via automatic differentiation. 

To enhance the model performance, we  introduce the \textit{curriculum} training \citep{krishnapriyan2021characterizing}  and \textit{causal} training \citep{wang2022respecting} in the training process.

\textbf{\textit{Curriculum} training}  starts with a simple PDE system and progressively solves the target PDE system. For this example, it is accomplished by 
minimizing the above PINN loss with a lower advection coefficient $c=10$  first and then gradually  increasing $c$ to the target value (i.e. $c=50$) during training. 

\textbf{\textit{Causal} training}  aims to impose temporal causality during the training of a PINNs model by appropriately re-weighting the PDE residual loss
at each iteration of gradient descent. Specifically, we split the 
temporal domain into $M$ chunks $[0, \Delta t], [\Delta t, 2 \Delta t], \dots$, and assign a weight to the corresponding temporal residuals losses $\mathcal{L}_r^i(\bm{\theta})$ as
\begin{align}
    \mathcal{L}_r(\bm{\theta}) = \sum_{i=1}^{M} w_i \mathcal{L}_i(\bm{\theta}),
\end{align}
with $w_1 = 1$, and
\begin{align}
    w_i = \exp(- \epsilon \sum_{k=1}^{i-1} \mathcal{L}_r^i(\bm{\theta})), \quad \text{for } i=2, \dots, M.
\end{align}
Here $\epsilon$ is a so-called \textit{causal} parameter, which is a user-specified hyper-parameter that determines the slope of the causal weights. We take $M=16$ and $\epsilon =0.1$ in this example.

\begin{figure}[H]
    \centering
    \includegraphics[width=0.9\textwidth]{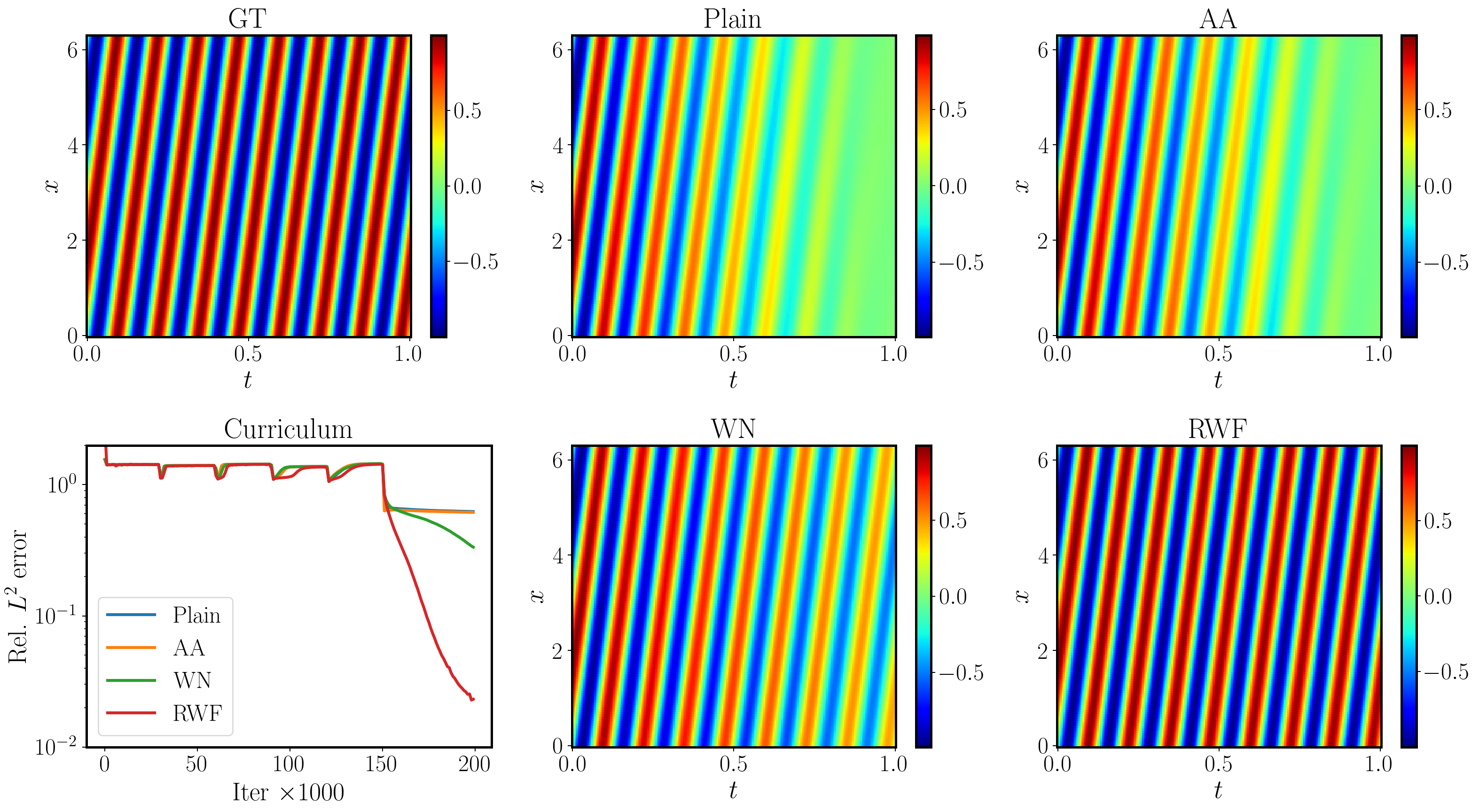}
       \caption{{\em Adection:} Predicted solutions of training PINNs with different weight parameterizations using curriculum training, as well as the evolution of the associated relative $L^2$ errors during training. }
    \label{fig: curriculum_pinn_adv}
\end{figure}

\begin{figure}[H]
    \centering
    \includegraphics[width=0.9\textwidth]{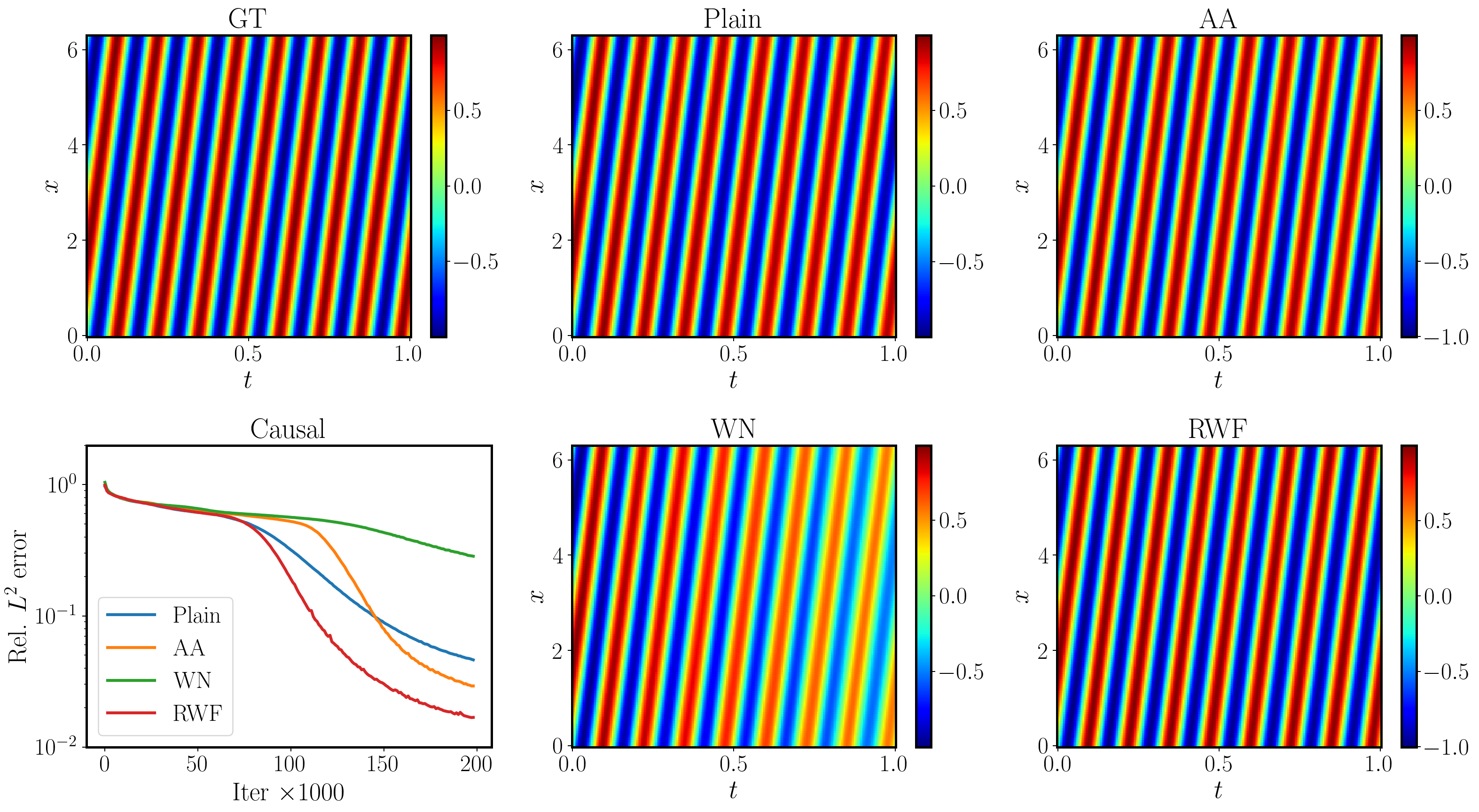}
     \caption{{\em Adection:} Predicted solutions of training PINNs with different weight parameterizations using causal training, as well as the evolution of the associated relative $L^2$ errors during training.}
    \label{fig: causal_pinn_adv}
\end{figure}

We initialize MLPs (5 layers, 256 channels, tanh activations)  with different weight parameterizations, and train each model with different strategies  for $2 \times 10^5$ iterations using the Adam optimizer \citep{kingma2014adam} with a start learning rate of  $10^{-3}$ and an exponential decay by a factor of $0.9$ for every $5,000$ steps. The resulting relative $L^2$ errors are presented in Table \ref{tab: pinn_adv}. In contrast to the failure of the other three parameterizations, RWF is the only one that enables PINN models to solve the advection equation with a reasonable and stable predictive accuracy. Further improvements can be obtained by combining RWF with  \textit{curriculum} or \textit{causal} training strategies. These conclusions are further clarified by the visualizations in Figures \ref{fig: pinn_adv}, \ref{fig: curriculum_pinn_adv} and \ref{fig: causal_pinn_adv}, where the predicted solutions corresponding to RWF are in excellent agreement with the ground truth.


\begin{table}[H]
    \centering
    \renewcommand{\arraystretch}{1.2}
    \begin{tabular}{c|c|c|c|c}
    \hline
        Advection     & Plain & AA  & WN & RWF (ours) \\
       \hline
       \hline
      Regular   & 28.82\% & 38.63\% & 46.34\% &  \textbf{4.14\%}\\
       Curriculum   & 62.37\% & 61.58\% & 33.41\% &  \textbf{2.31\%}\\
    Causal   & 4.61\% & 2.91\% & 28.48\% &  \textbf{1.67\%}\\
    \end{tabular}
    \caption{{\em Adection:} Relative $L^2$ errors of training PINNs with different weight parameterizations and training strategies.}
    \label{tab: pinn_adv}
\end{table}

\subsection{Navier-Stokes equation}

The underlying PDE system for this benchmark takes the form 
\begin{align}
    \bm{u} \cdot \nabla \bm{u}+\nabla p-\frac{1}{R e} \Delta \bm{u}&=0, \quad  (x,y) \in (0,1)^2, \\
    \nabla \cdot \bm{u}&=0, \quad  (x,y) \in (0,1)^2, \\
    \bm{u}&=(1,0), \quad    (x,y)  \text { on } \Gamma_1, \\
    \bm{u}&=(0,0), \quad  (x,y)  \text { on } \Gamma_2.
\end{align}
Here $\Gamma_1$ is the top boundary of a square cavity, while $\Gamma_2$ denotes the other three
sides of the cavity. We represent the unknown solution $u,v,p$ using a neural network $\bm{u}_{\bm{\theta}}$:
\begin{align}
    [x, y] \xrightarrow{\bm{u}_{\bm{\theta}}} [u_{\bm{\theta}}, v_{\bm{\theta}}, p_{\bm{\theta}}].
\end{align}
Then, the PDE residuals are defined  by
\begin{align}
    \mathcal{R}_{\bm{\theta}}^{u} &=  u_{\bm{\theta}}  \frac{\partial u_{\bm{\theta}}}{\partial x} +  v_{\bm{\theta}}  \frac{\partial u_{\bm{\theta}}}{\partial y} +   \frac{\partial p_{\bm{\theta}}}{\partial x} - \frac{1}{\text{Re}} (   \frac{\partial^2 u_{\bm{\theta}}}{\partial x^2}  +  \frac{\partial^2 u_{\bm{\theta}}}{\partial y^2} ), \\
    \mathcal{R}_{\bm{\theta}}^{v} &=   u_{\bm{\theta}}  \frac{\partial v_{\bm{\theta}}}{\partial x} +  u_{\bm{\theta}}  \frac{\partial v_{\bm{\theta}}}{\partial y} + \frac{\partial p_{\bm{\theta}}}{\partial y}  - \frac{1}{\text{Re}} (   \frac{\partial^2 u_{\bm{\theta}}}{\partial x^2}  +  \frac{\partial^2 u_{\bm{\theta}}}{\partial y^2} ), \\
    \mathcal{R}_{\bm{\theta}}^{c} &= \frac{\partial u_{\bm{\theta}}}{\partial x} + \frac{\partial v_{\bm{\theta}}}{\partial y}.
\end{align}
Given these residuals, along with a set of appropriate boundary conditions,  we can now formulate a loss function for training a physics-informed neural network as
\begin{align}
    \mathcal{L}(\theta)= \lambda_u \mathcal{L}_{u}(\theta)+ \lambda_v \mathcal{L}_{v}(\theta) + \lambda_{r_u} \mathcal{L}_{r_u}(\theta)+ \lambda_{r_v} \mathcal{L}_{r_v}(\theta)+ \lambda_{r_c} \mathcal{L}_{r_c}(\theta),
\end{align}
with
\begin{align}
    &\mathcal{L}_{u_b}(\theta)=\frac{1}{N_b} \sum_{i=1}^{N_b}\left[u\left(x_b^i, y_b^i\right)-u_b^i\right]^2,\\
    &\mathcal{L}_{v_b}(\theta)=\frac{1}{N_b} \sum_{i=1}^{N_b}\left[v\left(x_b^i, y_b^i\right)-v_b^i\right]^2, \\
    &\mathcal{L}_{r_u}(\theta)=\frac{1}{N_r} \sum_{i=1}^{N_r}\left[R_{\bm{\theta}}^u\left(x_r^i, y_r^i\right)\right]^2, \\
    &\mathcal{L}_{r_v}(\theta)=\frac{1}{N_r} \sum_{i=1}^{N_r}\left[R_{\bm{\theta}}^v \left(x_r^i, y_r^i\right)\right]^2, \\
    &\mathcal{L}_{r_c}(\theta)=\frac{1}{N_r} \sum_{i=1}^{N_r}\left[R_{\bm{\theta}}^c \left(x_r^i, y_r^i\right)\right]^2, 
\end{align}
where $\left\{\left(x_b^i, y_b^i\right), u_b^i\right\}_{i=1}^{N_b}$ and $\left\{\left(x_b^i, y_b^i\right), v_b^i\right\}_{i=1}^{N_b}$  denote the boundary data for the two velocity components at the domain boundaries $\Gamma_1$ and $\Gamma_2$, respectively, while $\left\{\left(x_r^i, y_r^i\right)\right\}_{i=1}^{N_r}$ is a set of collocation points for enforcing the PDE constraints. All of them are sampled randomly at each iteration of gradient descent. In experiments, we set $N_b = 256, N_r = 1024$ and $\lambda_u = \lambda_v = 100, \lambda_{r_u}= \lambda_{r_v}= \lambda_{r_c}= 1$.

\begin{figure}[H]
    \centering
    \includegraphics[width=0.9\textwidth]{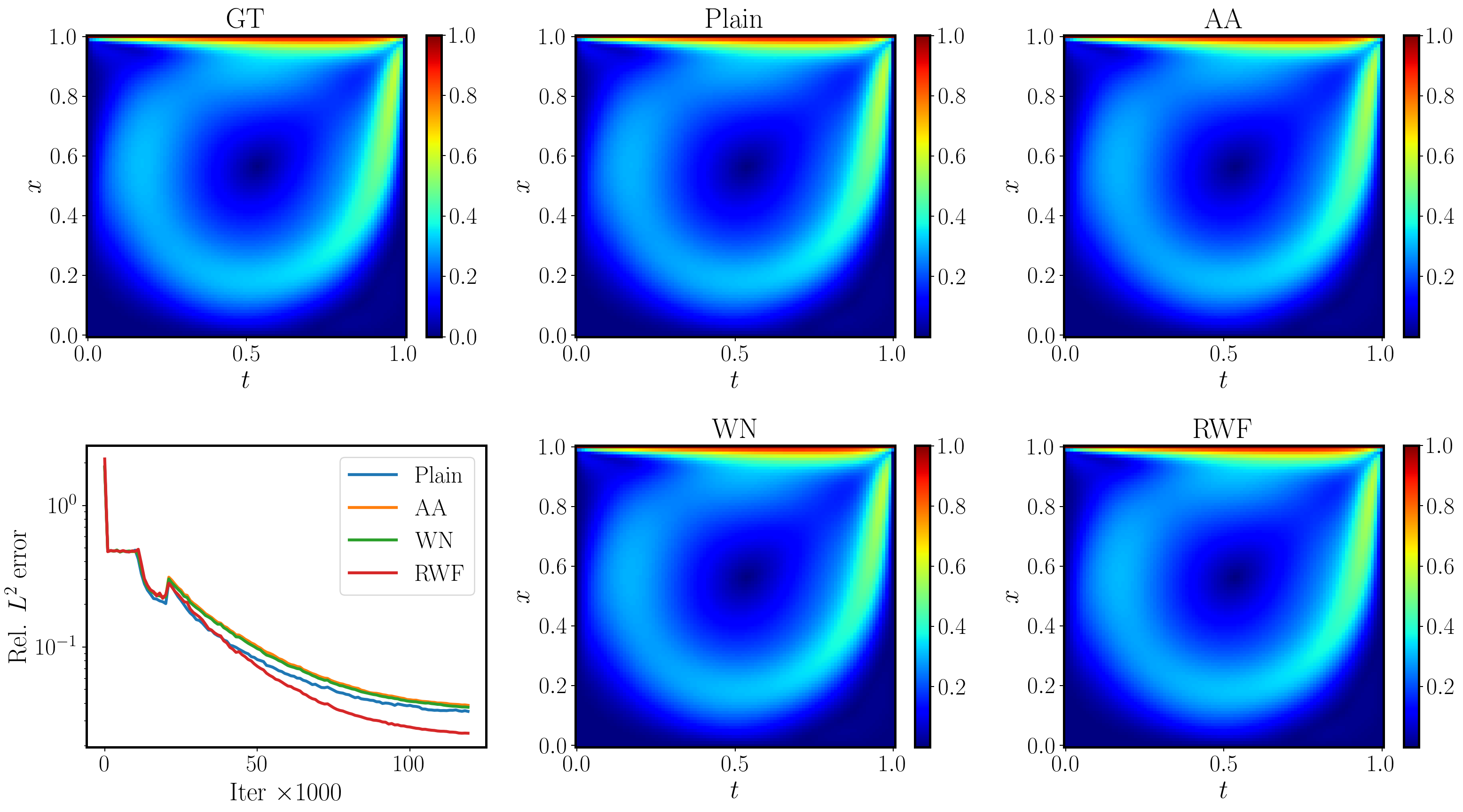}
      \caption{{\em Naiver-Stokes:} Predicted solutions obtained by modified MLPs with different weight parameterization, as well as the evolution of the associated relative $L^2$ errors during training.}
    \label{fig: modified_pinn_ns}
\end{figure}

We employ an MLP (5 layers, 128 channels, tanh activations)  to represent the latent variables of interest, and  train the network with different parameterizations for $10^5$ iterations using the Adam optimizer \citep{kingma2014adam} with a start learning rate of  $10^{-3}$ and an exponential decay by a factor of 0.9 for every $2,000$ training iterations. Moreover, we use a modified MLP architecure (see definition below) and curriculum training to enhance the model stability and performance. For the curriculum training, we minimize the loss with $Re = 100$ and $Re = 500$ for $2 \times 10^4$ iterations sequentially and change the Reynolds number to $Re = 1,000$  for the rest of the training.
The resulting relative $L^2$ errors are reported in Table \ref{tab: pinn_ns}. We can see that RWF performs the best among all parameterizations by a large margin. Some visualizations are shown in Figure \ref{fig: pinn_ns} and Figure \ref{fig: modified_pinn_ns}. We attribute this significant performance improvements to the fact that PINN models often suffer from poor initializations, and RWF precisely mitigates this by being able to reach better local minima that are located further away from the model initialization neighborhood.
 
\begin{table}[H]
    \centering
    \renewcommand{\arraystretch}{1.2}
    \begin{tabular}{c|c|c|c|c}
    \hline
        Navier-Stokes     & Plain & AA  & WN & RWF (ours) \\
       \hline
       \hline
      MLP    & 39.25\%  &  34.09\% & 30.98\% & \textbf{6.67\%} \\
       Modified MLP   & 3.51\%  &  3.87\% & 3.75\% & \textbf{2.46\%} \\
    
    \end{tabular}
    \caption{{\em Naiver-Stokes:} Relative $L^2$ errors of training conventional MLPs and modified MLPs with different weight parameterizations, respectively.}
    \label{tab: pinn_ns}
\end{table}

\paragraph{Modified MLP:}  In \citep{wang2021understanding} Wang {\em et al.} proposed a novel architecture that was demonstrated to outperform conventional MLPs across a variety of PINNs benchmarks. Here, we will refer to this architecture as "modified MLP". The forward pass of  a $L$-layer modified MLP is defined as follows
\begin{align}
    &\bm{U} = \sigma(   \bm{W}_1 \bm{x} + \bm{b}_1), \ \  \bm{V} = \sigma( \bm{W}_2 \bm{x}  + \bm{b}_2), \\
    &\bm{H}^{(1)} = \sigma( \bm{W}^{(1)}  \bm{x}+ \bm{b}^{(1)}), \\
    &\bm{Z}^{(l)} = \sigma(\bm{W}^{(l+1)} \bm{H}^{(k)} + \bm{b}^{(l+1)}), \ \ l=1, \dots, L - 1,\\
    &\bm{H}^{(l+1)} = (1 - \bm{Z}^{(l)}) \odot \bm{U}  +  \bm{Z}^{(l)}  \odot \bm{V}, \ \  l=1, \dots, L - 1, \\
   & \bm{u}_{\bm{\theta}}(\bm{x}) = \bm{W}^{(L+1)}  \bm{H}^{(L)} + \bm{b}^{(L+1)},
\end{align}
where $\sigma$ denotes a nonlinear activation function, $\odot$ denotes a point-wise multiplication.  All trainable parameters are given by
\begin{align}
    \bm{\theta} = \{\bm{W}_1, \bm{b}_1, \bm{W}_2, \bm{b}_1, (\bm{W}^{(l)}, \bm{b}^{(l)})_{l=1}^{L+1}\}.
\end{align}
This architecture is almost the same as a standard MLP network, with the addition of two encoders and a minor modification in the forward pass.  Specifically, the inputs $\bm{x}$ are embedded into a feature space via two encoders $\bm{U}, \bm{V}$, respectively, and merged in each hidden layer of a standard MLP  using a point-wise multiplication.

\section{Learning operators}
\label{appendix: deeponet}

\paragraph{Overview of DeepONets:} DeepONet is supposed to approximate an operator $G$ between functional spaces.  DeepONet consists of two separate neural networks referred to as the "branch net" and "trunk net", respectively. The branch net takes a vector $\bm{a}$ as input and returns a features embedding $[b_1, b_2,\dots, b_q]^T \in \R^q$ as output, where $\bm{a} = [\bm{a}(\bm{x}_1), \bm{a}(\bm{x}_2), \dots, \bm{a}(\bm{x}_m) ]$ represents a function $a$ evaluated at a collection of fixed locations $\{\bm{x}_i\}_{i=1}^m$.  The trunk net takes the continuous coordinates $\bm{y}$ as inputs, and outputs a features embedding $[t_1, t_2,\dots, t_q]^T \in \R^q$. The DeepONet output is obtained by merging the outputs of the branch and trunk networks together via a dot product
\begin{align}
    G_{\bm{\theta}}(\bm{a})(\bm{y}) = \sum_{k=1}^{q} \underbrace{b_{k}\left(\bm{a}\left(\bm{x}_{1}\right), \bm{a}\left(\bm{x}_{2}\right), \ldots, \bm{a}\left(\bm{x}_{m}\right)\right)}_{\text {branch }} \underbrace{t_{k}(\bm{y})}_{\text {trunk }},
\end{align}
where $\bm{\theta}$ denotes the collection of all trainable weight and bias parameters in the branch and trunk networks. These parameters can be optimized by minimizing the following mean square error loss
\begin{align}
\label{eq: loss_operator}
    \mathcal{L}(\bm{\theta}) &= \frac{1}{NP} \sum_{i=1}^N \sum_{j=1}^P \left|G_{\bm{\theta}} (\bm{a}^{(i)})(\bm{y}^{(i)}_j)- G (\bm{a}^{(i)})(\bm{y}^{(i)}_j) \right|^2  \\
    &= \frac{1}{NP} \sum_{i=1}^N \sum_{j=1}^P \left| \sum_{k=1}^q b_k(\bm{a}^{(i)}(\bm{x}_1), \dots, \bm{a}^{(i)}(\bm{x}_m))t_k(\bm{y}^{(i)}_j)  - G (\bm{a}^{(i)})(\bm{y}^{(i)}_j) \right|^2,
\end{align}
where $\{ \bm{a}^{(i)} \}_{i=1}^N$ denotes $N$ separate input functions sampled from a function space $\mathcal{U}$. For each  $\bm{a}^{(i)}$, $\{\bm{y}^{(i)}_j\}_{j=1}^P$ are $P$ locations in the domain of $G(\bm{a}^{(i)})$, and $G(\bm{a}^{(i)})(\bm{y}_j^{(i)})$ is the corresponding output data evaluated at $\bm{y}_j^{(i)}$ .  Contrary to the fixed sensor locations of $\{x_i\}_{i=1}^m$, we remark that the locations of $\{\bm{y}^{(i)}\}_{j=1}^P$ may vary for different $i$, thus allowing us to construct a continuous representation of the output function $G(a)$.

\textbf{Remark:} All parameterizations (AA, WN, RWF) are applied to every dense layer of the DeepONet architecture (in the cases where such parametrizations are employed).

\begin{table}[H]
    \centering
       \renewcommand{\arraystretch}{1.2}
    \begin{tabular}{c|c|c|c|c}
      Case   & Plain & AA & WN & RWF \\
      \hline
     DR & $1.09\% \pm 0.53\%$ &  $0.95\% \pm 0.46\%$ & $0.97\%  \pm 0.46\%$ & $\mathbf{0.50\% \pm 0.26\%}$  \\
     Darcy & $2.03\% \pm 1.54\%$ &  $2.06\% \pm 1.70 \%$ & $2.05\% \pm 2.00\%$ & $\mathbf{1.67\% \pm 1.70\%}$  \\
     Burgers &  $5.11\% \pm 3.79\%$  &  $4.71\% \pm 3.39\%$  &  $ 4.37\% \pm 2.63\%$ &  $\mathbf{2.46\% \pm 2.05\%}$  \\
    \end{tabular}
    \caption{{\em Learning operators:} Relative $L^2$ errors of trained (physics-informed) DeepONets over the test data-set of different examples.}
    \label{tab: deeponet_l2_error}
\end{table}

\subsection{Diffusion-reaction}
The underlying PDE for this benchmark takes the form 
\begin{align}
    \frac{\partial u}{\partial t}= D \frac{\partial^{2} u}{\partial x^{2}}+k u^{2}+a(x), \quad (x,t) \in (0,1) \times (0,1],
\end{align}
subject to zero initial and boundary conditions.

\paragraph{Data Generation:} We  sample $N = 5,000$ input functions $a(x)$ from a GRF with length scale $l=0.2$ and solve the diffusion-reaction system using a second-order implicit finite difference method on a $100 \times 100$ equispaced grid. To generate the training data, we randomly take $P=100$ measurements from each solution. The test data-set contains another 100 solutions evaluated at the same mesh.

\begin{figure}[H]
    \centering
    \includegraphics[width=0.9\textwidth]{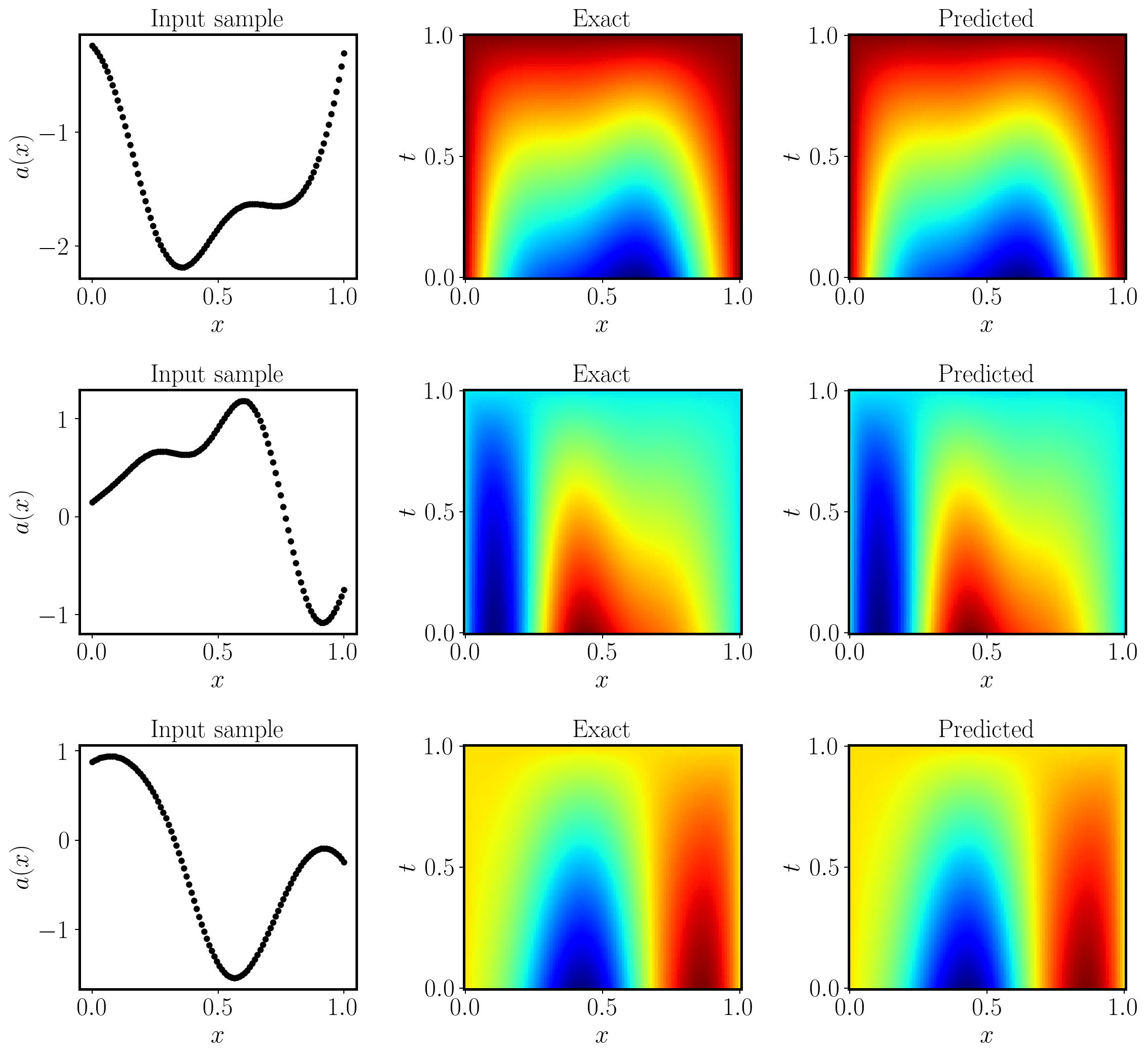}
    \caption{{\em Diffusion-reaction (DR) PDE:} Predicted solutions of a trained DeepONet with random weight factorization, corresponding to randomly chosen input samples in the test data-set.}
    \label{fig: dr_pred}
\end{figure}

We represent the solution operator by a DeepONet  $G_{\bm{\theta}}$ , where the branch and trunk networks are two separate MLPs (5 layers, 64 channels, tanh activations). The model is trained  for $5 \times 10^4$ iterations using the Adam optimizer \citep{kingma2014adam} with a start learning rate of $10^{-3}$ and an exponential decay by a factor of $0.9$ for every $1000$ steps. The mean and standard deviation of the relative $L^2$ errors over the test date-set are shown in Table  \ref{tab: deeponet_l2_error}. Figure \ref{fig: dr_pred} provides several representative predictions using RWF.

\subsection{Darcy flow}
The PDE system for this benchmark takes the form
\begin{align}
-\nabla \cdot(a \cdot \nabla u) &=1, \quad (x,y) \in(0,1)^{2}, \\
u &=0, \quad    (x,y) \in \partial(0,1)^{2}.
\end{align}

\paragraph{Data Generation:} We sample the coefficient function $a$ from a Gaussian random field with a length scale $l = -0.5$ and solve the associated Darcy flow using finite element method. The training data contains $2,000$ solutions evaluated at a  $64 \times 64$ uniform mesh while the test data contains $100$ solutions on the same mesh.

\begin{figure}[H]
    \centering
    \includegraphics[width=0.9\textwidth]{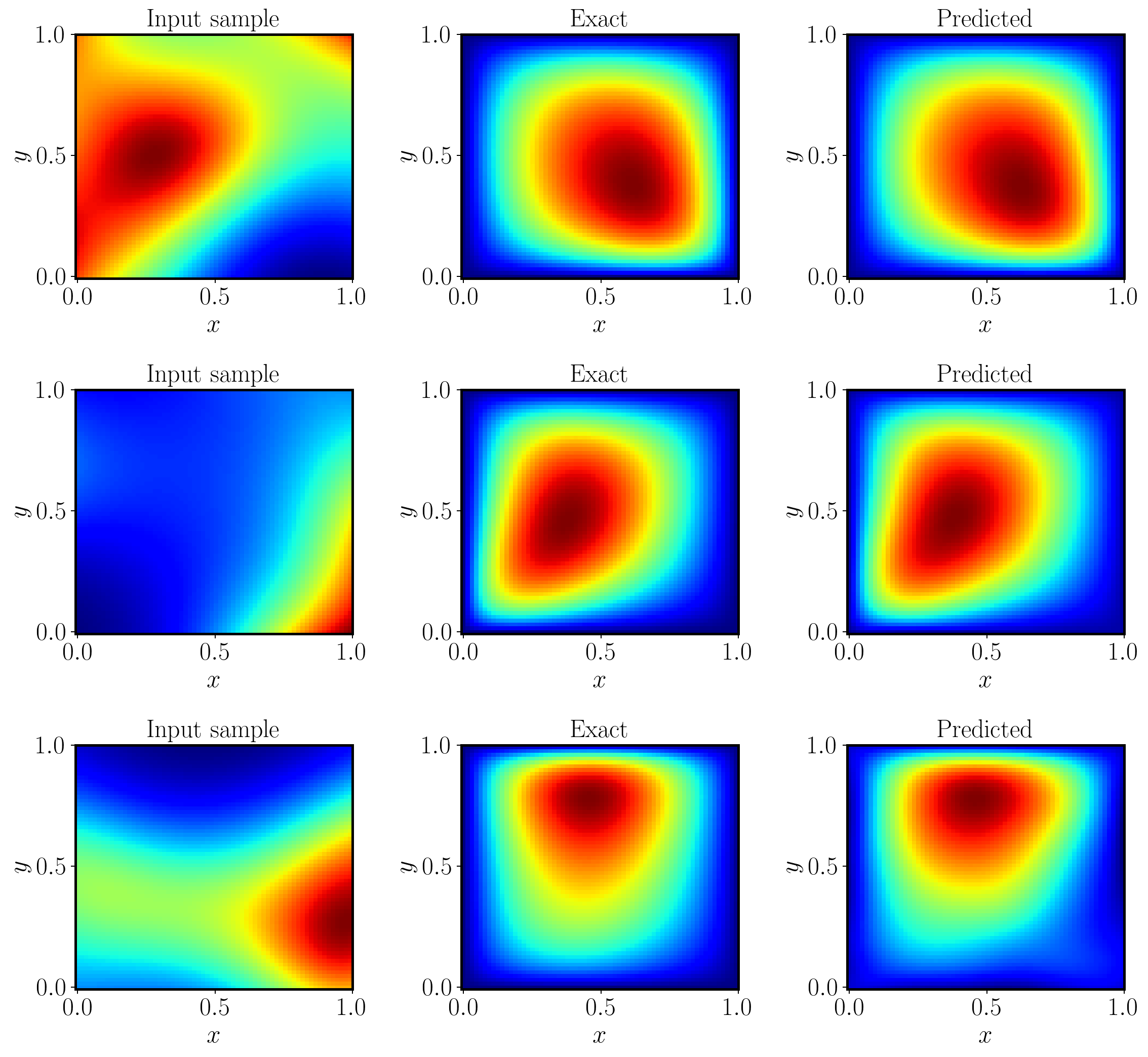}
     \caption{{\em Darcy PDE:} Predicted solutions of a trained DeepONet with random weight factorization, corresponding to randomly chosen input samples in the test data-set.}
    \label{fig: darcy_pred}
\end{figure}

We represent the solution operator by a DeepONet  $G_{\bm{\theta}}$, where the branch network is a convolutional neural network (CNN) for extracting latent feature representation of the input coefficients and the trunk network is a 4-layer MLP with GELU activations and $128$ neurons per hidden layer.  We train each model with different parameterizations  for $5 \times 10^4$ iterations using the Adam optimizer \citep{kingma2014adam} with a start learning rate of $10^{-3}$ and an exponential decay by a factor of $0.9$ for every $1000$ steps. The results are summarized in Table \ref{tab: deeponet_l2_error} and some   predicted solutions are plotted in Figure \ref{fig: darcy_pred}.

\subsection{Burgers' equation}

Recall the one-dimensional Burgers' equation is given by
\begin{align}
    \label{eq: Burger_eq}
    \frac{du}{dt} + u \frac{du}{dx} - \nu \frac{d^2 u}{dx^2} = 0, \quad (x, t) \in (0,1) \times (0,1], 
\end{align}
subject to initial and the periodic boundary conditions
\begin{align}
        u(x,0) &= a(x), \quad x \in (0,1), 
\end{align}

\paragraph{Data Generation:} The training data only consists of $1,000$ input functions $ a$ sampled from  a Gaussian random field $\sim \mathcal{N}\left(0,  25^2(-\Delta+5^2 I)^{-4}\right)$. 
To generate the test data-set, we sample another $100$  input functions from the same Gaussian random field and solve the Burgers' equation  using 
the Chebfun package \cite{driscoll2014chebfun} with a spectral Fourier discretization and a fourth-order stiff time-stepping scheme (ETDRK4) \cite{cox2002exponential} with a time-step size of $10^{-4}$.  Temporal snapshots of the solution are saved every $\Delta t = 0.01$ to give us 101 snapshots in total.  Consequently, the test data-set contains $500$ realizations evaluated at a $100 \times 101$ spatio-temporal grid.

Our objective here is to learn the solution operator mapping initial conditions $a(x)$ to the associated full spatio-temporal solution $u(x,t)$. Here proceed by representing  the solution operator by a modified DeepONet architecture \cite{wang2021improved} outlined below.
To impose the periodic the exact boundary condition, we apply a Fourier feature mapping  to the input coordinates before passing them through the trunk network
\begin{align}
    [x, t] \rightarrow [\cos(2 \pi x), \sin(2 \pi x), t].
\end{align}

\begin{figure}[H]
    \centering
    \includegraphics[width=0.9\textwidth]{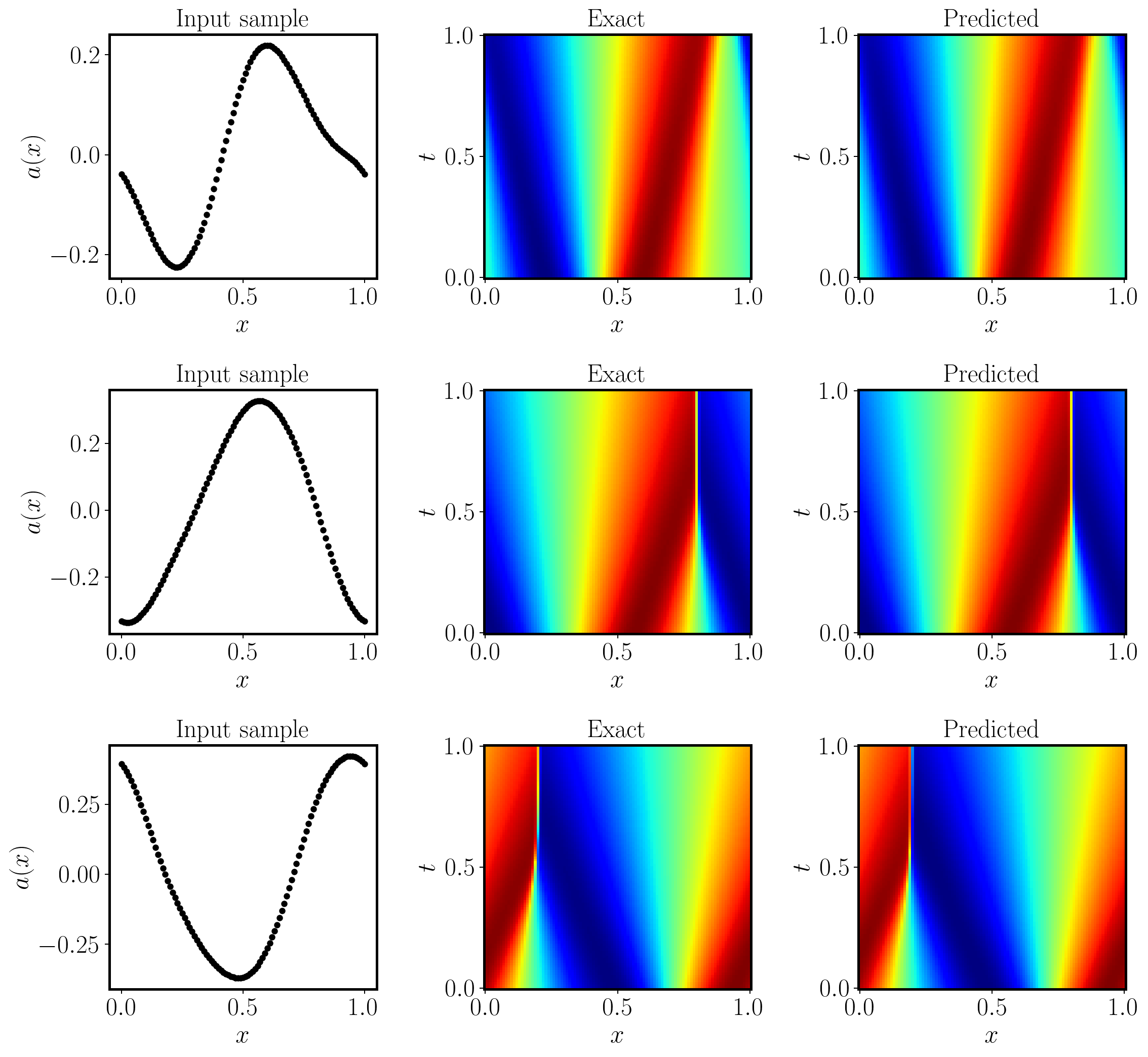}
         \caption{{\em Burgers PDE:} Predicted solutions of a trained physics-informed DeepONet with random weight factorization, corresponding to randomly chosen input samples in the test data-set.}
    \label{fig: burger_pred}
\end{figure}

Then the PDE residual is then defined by
\begin{align}
    \label{eq: Burger_residual}
    R_{\bm{\theta}}[a] = \frac{\partial G_{\bm{\theta}}(\bm{a})}{\partial t} + G_{\bm{\theta}}(\bm{a})\frac{\partial G_{\bm{\theta}}(\bm{a})}{\partial x} 
    - \nu \frac{\partial^2 G_{\bm{\theta}}(\bm{a})}{\partial x^2},
\end{align}
Consequently, a physics-informed DeepONet can be trained by minimizing the following weighted loss function 
\begin{align}
    \label{eq: Burger_loss}
    \mathcal{L}(\bm{\theta}) = \lambda_{ic}  \mathcal{L}_{ic}(\bm{\theta})  + \lambda_r \mathcal{L}_{r}(\bm{\theta}),
\end{align}
where 
\begin{align}
     \mathcal{L}_{ic}(\bm{\theta}) &= \frac{1}{NP}\sum_{i=1}^N \sum_{j=1}^{P} \left|G_{\bm{\theta}}(\bm{a}^{(i)})(x^{(i)}_{ic, j}, 0 ) - u^{(i)}(x^{(i)}_{ic, j})  \right|^2, \\
    \mathcal{L}_{r}(\bm{\theta}) &= \frac{1}{NQ}\sum_{i=1}^N \sum_{j=1}^{Q} 
    \left|    R_{\bm{\theta}}^{(i)}(x^{(i)}_{r,j}, t^{(i)}_{r,j})   \right|^2.
\end{align}
For this example, we take $N = 64, P=100$ and $Q = 512$, which means that we randomly sample $N=64$ input functions from the training data-set and $Q = 512$ collocation points inside the computational domain. In particular, we set $\lambda_{ic} = 100, \lambda_{r} = 1$ for better enforcing the initial condition across different input samples. The model with different parameterizations is trained for
$10^5$ iterations using the the Adam optimizer \citep{kingma2014adam} with a start learning rate of $10^{-3}$ and an exponential decay by a factor of $0.9$ for every $2,000$ steps. We report the test errors in Table \ref{tab: deeponet_l2_error} and visualize some predicted solutions in Figure \ref{fig: burger_pred}.

\paragraph{Modified DeepONet:} \cite{wang2021improved} modify the forward pass of an L-layer  DeepONet as follows
\begin{align}
     & \bm{U} = \phi( \bm{W}_a \bm{a} + \bm{b}_a), \ \  \bm{V} = \phi( \bm{W}_y \bm{y} + \bm{b}_y), \\
     & \bm{H}_a^{(1)} = \phi(\bm{W}^{(1)}_a \bm{a}  + \bm{b}^{(1)}_a), \ \ \bm{H}_y^{(1)} = \phi( \bm{W}^{(1)}_y \bm{y} + \bm{b}^{(1)}_y), \\
    & \bm{Z}_a^{(l)} = \phi(\bm{W}^{(l)}_a \bm{H}^{(l)}_a  + \bm{b}^{(l)}_a), \ \ \bm{Z}_y^{(l)} = \phi( \bm{W}^{(l)}_y \bm{H}^{(l)}_y + \bm{b}^{(l)}_y), \quad l = 1, 2, \dots, L-1, \\
    \label{eq: arch_embed_1}
    & \bm{H}^{(l+1)}_a = (1 - \bm{Z}^{(l)}_a) \odot \bm{U}  +  \bm{Z}^{(l)}_a  \odot \bm{V}, \quad l = 1, \dots, L-1, \\
     \label{eq: arch_embed_2}
    & \bm{H}^{(l+1)}_y = (1 - \bm{Z}^{(l)}_y) \odot \bm{U}  +  \bm{Z}^{(l)}_y  \odot \bm{V}, \quad l = 1, \dots, L-1, \\
    & \bm{H}_a^{(L)} = \phi(\bm{W}^{(L)}_a \bm{H}^{(L-1)}_a  + \bm{b}^{(L)}_a), \ \ \bm{H}_y^{(L)} = \phi( \bm{W}^{(L)}_y \bm{H}^{(L-1)}_y   + \bm{b}^{(L)}_y), \\
    &G_{\bm{\theta}}(\bm{a})(\bm{y}) = \left\langle \bm{H}_a^{(L)}, \bm{H}_y^{(L)}        \right\rangle,
\end{align}
where $\odot$ denotes point-wise multiplication, $\phi$ denotes a activation function, and  $\bm{\theta}$ represents all  trainable parameters of the DeepONet model. In particular, $\{\bm{W}_a^{(l)},  \bm{b}_a^{(l+1)} \}_{l=1}^{L+1}$ and $ \{\bm{W}_y^{(l)},  \bm{b}_y^{(l+1)} \}_{l=1}^{L+1} $ are the weights and biases of the branch and trunk networks, respectively. 
we embed the DeepONet inputs $\bm{a}$ and $\bm{y}$ into a high-dimensional feature space via two encoders $\bm{U}, \bm{V}$, respectively. Instead of just merging the propagated information in the output layer of the branch and trunk networks, we merge the embeddings $\bm{U}, \bm{V}$ in each hidden layer of these two sub-networks using  a point-wise multiplication (equation (\ref{eq: arch_embed_1}) - (\ref{eq: arch_embed_2})). Heuristically, this design may not only help input signals propagate through the DeepONet, but also enhance its capability of representing non-linearity due to the extensive use of point-wise multiplications.

\end{document}